\documentclass[10pt,twocolumn,letterpaper]{article}

\usepackage{cvpr} 
\usepackage{times}
\usepackage{epsfig}
\usepackage{graphicx}
\usepackage{amsmath}
\usepackage{amssymb}
\usepackage{booktabs}
\usepackage{threeparttable}
\usepackage{caption}
\usepackage{enumitem}
\usepackage{graphicx}
\usepackage{makecell}
\usepackage{hyperref}
\usepackage{verbatim}

\begin{document}

\title{Intentional Electromagnetic Interference Attacks on Facial Recognition}

\author{Tyler Fitzsimmons$^{\dag,\ddag}$ and Adam Czajka$^\dag$\\
$^\dag$Department of Computer Science and Engineering\\
University of Notre Dame, Notre Dame, IN 46556, USA\\
$^\ddag$Naval Surface Warfare Center, Crane Division, Crane, IN 47522, USA
\\
{\small\tt\{tfitzsi3,aczajka\}@nd.edu}
}

\maketitle
\thispagestyle{empty}

\begin{abstract}
Attacks on general computer vision algorithms are often relegated to the digital domain, with the optimization performed purely in the digital world and then translated to physical mediums for implementation. 
In the field of biometrics, including facial recognition, physical presentation attacks targeting biometric sensors are dominant and present significant opportunity and risk. This paper highlights a critical vulnerability in the physical-to-digital pipeline of biometric sensors and provides a standardized approach for testing facial recognition system robustness against hardware attacks, going beyond and potentially complementing presentation attacks (as defined in ISO/IEC 30107 standard series). Specifically, in this work we (a) demonstrate that intentional electromagnetic interference is possible to be conducted with commonly accessible radio frequency (RF) equipment, (b) assess the robustness of state-of-the-art face recognition methods against RF-based attacks, and (c) provide a dataset composed of face images captured with and without electromagnetic interference to serve as a new benchmark for testing modern face matchers against RF-sourced interference. 
\end{abstract}

\section{Introduction}
As deep learning-based facial recognition systems become ubiquitous in secure access control, understanding their vulnerability to physical-layer perturbations is critical. While digital adversarial machine learning, as well as ISO/IEC 30107-defined presentation attacks and presentation attack detection (PAD) methods are well-documented (often utilizing optimized datasets in a digital paradigm, or focusing on non-compliant presentations to the sensor) these existing approaches frequently omit the hardware-level vulnerabilities of the camera itself and typically assume gray-box or white-box access to the victim model \cite{Akhtar2018Survey,Yu_TPAMI_2022}.

This paper addresses these limitations by investigating the impact of hardware-level intentional electromagnetic interference (IEMI) on the reliability of facial recognition systems in a black-box setting. Specifically, we examine the susceptibility of common and state-of-the-art face matchers (including VGG-Face \cite{Caoetal2018}, SFace \cite{Zhongetal2021}, GhostFaceNet \cite{Alansarietal2023}, buffalo\_l \cite{insightface}, antelopv2 \cite{insightface}, VeriLook \cite{verilook}, and DINOv3 \cite{dinov3_2025}\cite{dinov3_hf_model}) to targeted electric and magnetic field interference. By quantifying the False Non-Match Rate (FNMR) induced by radio frequency (RF)-driven image artifacts, our results demonstrate that such interference causes significant attack success. Specifically, the results and analysis presented in this paper reveal that at an False Match Rate (FMR) between 0.1-5\%, the targeted RF attack increases embedding distances to levels where the FNMR rises to 100\% 
effectively creating a reversible, non-obvious, physical attack for real-world applications. 

\subsection{Contributions}
\noindent
We summarize our contributions as: 
\begin{enumerate}[nosep]
    \item A new physical attack leveraging low-cost, untargeted, black-box hardware used with face recognition. 
    \item An evaluation of how this attack is generalized across modern neural network-based backbones and loss functions used to train these models.  
    \item New dataset of face videos representing 50 identities, recaptured\footnote{The authors obtained the permission of the original dataset owners to share the re-captured clean and under-attack face videos.} by a camera under the IEMI attack along with recaptured videos without the IEMI attack, to serve as a new benchmark for testing reliability of face recognition models. 
    \item Method and source code\footnote{\url{https://github.com/CVRL/EMI-Attacks-Face}} generating IEMI-like pattern and overlaying it on the existing face images, which may serve as a useful IEMI-specific augmentation technique for training face recognition models.
\end{enumerate}

\begin{figure}[!htb]
     \includegraphics[width=\linewidth]{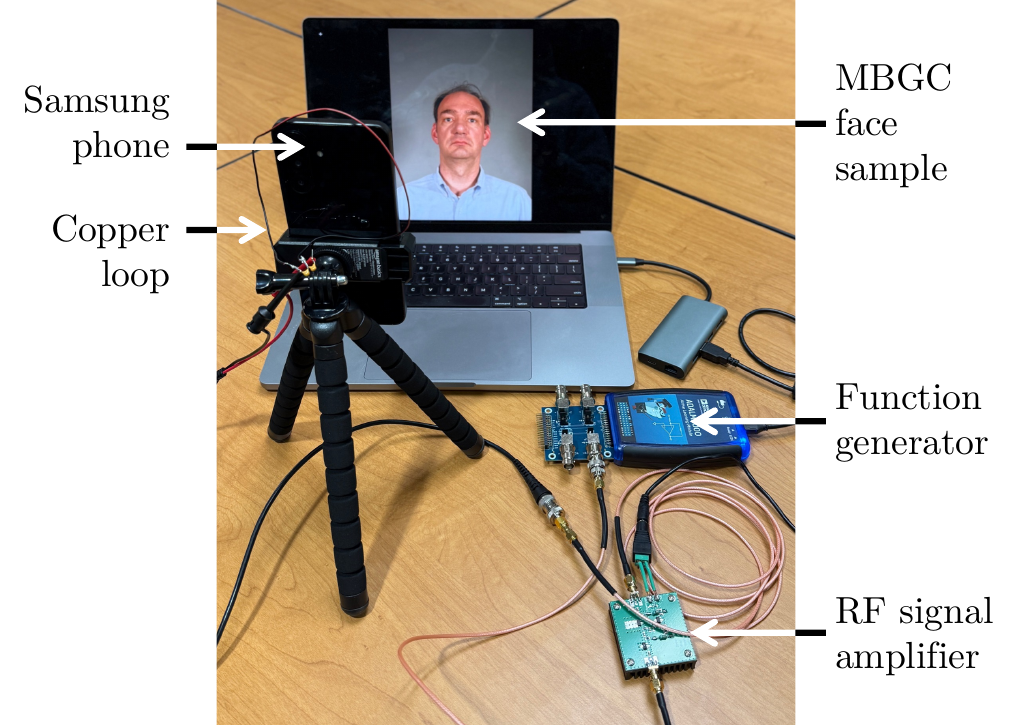}
     \caption{\textbf{Data capture setup:} The Samsung A26 front-facing camera is placed parallel to a MacBook Pro screen, collecting an  MBGC (Multiple Biometric Grand Challenge) V2 face sample. The function generator, with user-defined parameters, transmits through the RF signal amplifier for increased current, along the coaxial cable to the copper loop. The single turn copper loop is placed around the Samsung phone and generates an electromagnetic field, causing the attack due to the specific waveforms generated. }
     \label{fig:setup}
\end{figure}

\subsection{Summary of the Approach}
Since many current face recognition models (not explicitly equipped with presentation attack detection, which usually consists of separate models) accept face images recaptured from a computer screen, we built a test bed whose advantage is simplicity and ease of replication in any lab, without the need for time-consuming human subject collection. In this setup, the face images from a publicly-available face recognition benchmark are displayed on a high-quality standard screen and recaptured by the camera under test to provide ``clean'' and ``attack'' samples, as shown in Fig. \ref{fig:setup}.

Our approach is valid for many real-world facial recognition attack scenarios and the hardware setup can be highly mobile. An attack scenario where an adversary has physical access to a user's smartphone and desires to access their device is highly analogous to our experiment. If the phone has previously been exploited and the IEMI frequency is known, the emitter frequencies are quite deterministic. If a completely unknown device is encountered, the attacker can still sweep through various frequencies, especially for smartphones, where Focal Plane Array (FPA) designs are often similar and manufactured by few companies. The adversary simply needs to place the copper loop relatively close to the phone, where the center of the loop is ideally concentric with the phone's camera. 

Since smart phones typically only save one to two identities per device, the attack can be nuanced finding ideal attack frequencies. Another real-world scenario is to target cameras used for biometrics such as webcams or specialized hardware. These setups are the eyes for larger databases with significantly more identities, actually making the untargeted attack more likely to succeed, and realistic for office settings or residential security cameras. In addition to denying identification of existing identities, this method is discreet enough for an adversary to deploy to \textit{prevent} high-quality collection of their face, given access to a camera. This scenario is also realistic for government and commercial surveillance efforts. 

\section{Related Work}
\subsection{Physical Attacks on Computer Vision Sensors}
Recent research on imaging sensor security has transitioned from digital adversarial perturbations to complex physical-layer signal injections targeting the CMOS/CCD array, Optical Image Stabilization (OIS), and the Image Signal Processor (ISP). Goiffon \cite{goiffon2023} showed how radiation effects on CMOS detectors could be dialed in to affect a single pixel on an array. This novel contribution introduced the idea of controllable, IEMI effects on cutting edge sensors, specifically 
through exploiting the rolling shutter to negatively affect image classification models. 

Similar to other related works, but unlike our approach, the work by Sayles \etal \cite{sayles2021invisible} requires white box access to optimize the attack through the loss function. The attack, using light emitting diodes (LED), was up to 99\% successful for their target model. 
Ren \etal \cite{ren2025ghostshot} 
used IEMI to manipulate 
open-air CCDs without housings for easier coupling. Through testing of two models their results showed high success, with F-1 scores dropping by up to 71\% when attacked. Liu \etal \cite{liu2025} explored four types of attacks, a Cartesian product of (a) targeted and untargeted methods and (b) image classification and object detection tasks. Their approach leveraged five CCDs and accomplished up to 99\% attack success rate for four popular models (YOLOv3/4, Inceptionv3, ResNet101) \cite{bochkovskiy2020yolov4, he2016deep, redmon2018yolov3,  szegedy2016rethinking}. 

Ji \etal \cite{ji2021poltergeist} approached physical attacks in a slightly different way, but exploiting the OIS sensors in modern cameras with acoustic-based attacks, which aim to undo the benefit of OIS sensors and intentionally create negative affects performed in a black box setting. The authors 
were able 
to successfully create an attack up to 46.2\% of the time. 

Finally, Zhang \etal \cite{zhang2025rainbow} demonstrated rainbow effects on CCD and CMOS detectors with IEMI. Their approach was geared towards self-driving car applications and showed promise as an attack vector. Zhang \etal's paper is the only work where the attack was completely black box and non-optimized, which is likely why their results show increased precision scores under certain models attacked. 

Our work addresses current gaps and demonstrates the attack vector through closed hardware on a modern imaging device, at a single non-optimized frequency (black box), across multiple FR models.

\subsection{Physical Attacks on Facial Recognition Systems}
In addition to the broader adversarial machine learning) space, biometric presentation attacks and their detection is a very active and large research area also in face recognition, with a massive literature offering both systematic reviews of attacks as well as countermeasures. Presentation attack instruments include popular objects such as 3D masks \cite{liu2025}, and more sophisticated attacks including digitally created patches projected onto faces \cite{Nguyenetal2020} or adversarial glasses \cite{sharif2016}. This paper is not related to presentation attacks, as defined by ISO/IEC 30107, and thus we recommend several good survey papers summarizing various types of face presentation attacks and their evolution spanning non-conformant presentations of authentic faces, printouts, video replay, using accessories, occlusions, and masks \cite{Antil_Neurocomputing_2025,leyva2023,Ramachandra_CSUR_2017}.


\section{Approach}
\subsection{Face Image Samples}

The original (non-attack) face images were samples from the publicly available Multiple Biometric Grand Challenge (MBGC) Version 2 Dataset\footnote{\url{https://www.nist.gov/programs-projects/multiple-biometric-grand-challenge-mbgc}}. These photos representing 50 unique identities were displayed on a MacBook Pro laptop screen and recaptured by the tested phone camera, as presented in Fig. \ref{fig:setup}. We chose this face image re-capturing approach, rather than collecting data with human subjects, to introduce the same physical noise in all test scenarios independent of face expression and differences in presentations, and to make similar tests easy to be replicated in other labs. The fact of photographing the already-captured faces is not important from the PAD point of view, since we are not assessing the PAD performance. Further, both the \textit{clean} and \textit{attack} samples were recaptured using the MacBook Pro display to eliminate any screen specific artifacts biasing one dataset. 
     
\subsection{Hardware}
The experiments in this paper focused on a camera installed in Samsung A26 phones, as the target for all attacks. While there are four cameras on the phone, three rear and one front-facing camera, the focus of the attacks was on the front-facing camera as it is the primary facial recognition camera \cite{samsung_galaxy_a26}. Using the front facing camera, face pictures were presented to the camera under a static attack scenario (described below in Sec. \ref{sec:AttackScenario}) using IEMI to create the effects. 

This was accomplished by connecting an Arbitrary Function Generator (AFG, model Tektronix AFG31000 Series \cite{tek_afg31000}) to a copper coil (24 AWG 9cm diameter, single loop), resonant at the specific frequencies, to emanate enough electric and magnetic field to interfere with the camera. The AFG is connected to a 2 Watt RF amplifier (responsive from 1-930MHz \cite{rf_amp_1_930mhz}) for additional current. An important note in contrast with other prior work is that our target phone was {\bf not modified or opened in any way}. That is, the effects reported in this paper have been generated using a non-modified commercial-off-the-shelf device.

\begin{table}[htbp]
    \centering
        \caption{Estimated cost of the equipment used in the experiments. The authors used devices available in their lab. However, cheaper versions can be used, which would result in a total cost around 398 USD (350 USD for a function generator and 30 USD for a camera).}
        \label{tab:exp_hardware}\vskip2mm
        \begin{tabular}{l l} 
            \toprule
            \textbf{Item} & \textbf{Cost (USD)} \\ 
            \midrule
            Function Generator & 3,200 \\
            RF Amplifier       & 13 \\
            Copper Loop        & 5 \\
            Camera/Phone       & 200\tnote{*} \\
            \midrule 
            \textbf{Total}     & \textbf{3,418} \\ 
            \bottomrule
        \end{tabular}
\end{table}

\subsection{Attack Scenario}
\label{sec:AttackScenario}
To determine the static attack scenario, a set of experiments were performed to establish qualitative negative effects on the cameras ability to discern new faces and the overall maximization of camera distortion. 
The AFG baseline generator is capable of generating basic continuous functions, Amplitude/Frequency modulation (AM/FM), frequency sweeps, and bursts. Many of these capabilities were explored for a maximum negative effect on the camera, though this work specifically reports the effects of a single FM attack. The FM was experimentally varied across multiple carrier and modulation shapes, in addition to actual carrier and modulation frequencies. For the Samsung A26 phone, FM using a square carrier at 11.465 MHz with a triangle modulation at 190 kHz created noticeable distortions in captured images. The distortions were vertical lines with color changing properties, as seen in Fig. \ref{fig:four_photos}c. In addition, the vertical line width and horizontal scrolling of the lines varied with modulation frequency $\pm$1 kHz. The overall attack aimed to reduce the obvious nature of most adversarial attacks while maximizing attack success. The vertical lines present at the specific frequency modulation represent an often successful attack, yet are subtle to human observers. 

\subsection{Face Matchers Selected for Experiments}

\begin{table}[htbp]
    \centering
    \begin{threeparttable}
        \caption{Face Recognition Models Selected for Experiments.}
        \label{tab:Model_comparison}
        \begin{tabular}{l l p{1.8cm} p{2.2cm}} 
            \toprule
            \textbf{Model} & \textbf{Year} & \textbf{Loss} & \textbf{Backbone/ Training Dataset} \\
            \midrule
            {\it VGG-Face}     & 2015 & Softmax Triplet Loss  & VGG-16/ VGG-Face \\\hline
            {\it SFace}        & 2021 & Sigmoid Hypersphere   & ResNet-50/ CASIA-WebFace \\\hline
            {\it GhostFace}\\{\it Net} & 2022 & ArcFace     & Ghost Modules/ MS1MV3 \\\hline
            {\it buffalo\_l}   & 2023 & ArcFace       & ResNet-50/ MS1MV3 \\\hline
            {\it antelopev2}   & 2024 & ArcFace       & ViT/Glint360K \\\hline
            {\it VeriLook} & 2024   &  Proprietary  & Proprietary  \\\hline
            {\it DINOv3} & 2025 & Self-supervised     & ViT/LVD-1689M \\
            \bottomrule
        \end{tabular}
    \end{threeparttable}
\end{table}

All image pairs were compared using six open-source face recognition models ({\it VGG-Face} \cite{Caoetal2018}, {\it SFace} \cite{Zhongetal2021}, {\it GhostFaceNet} \cite{Alansarietal2023}, two {\it InsightFace} models ({\it buffalo\_l} and {\it antelopev2}) \cite{insightface}, and {\it DINOv3}-based approach \cite{dinov3_2025}) and one commercial face recognition method ({\it VeriLook} \cite{verilook}), all developed in the last decade. The methods represent diverse architectures and training paradigms, providing comprehensive evaluation across face recognition technology.

We selected {\it VGG-Face} because it is based on the VGG-16 architecture adapted for facial recognition, trained on 2.6 million images. While not as deep as more recent architectures, its widespread deployment and documentation make it a valuable baseline for attack evaluation. The model uses Euclidean distance for face matching and has been extensively studied in adversarial robustness literature. {\it VGG-Face} uses softmax loss for initial classification and triplet loss for fine-tuning. 

{\it SFace} uses a ResNet-50 backbone and addresses class-imbalance challenges by optimizing face embeddings on a unit hypersphere, ensuring better generalization across varying facial attributes. Considering smaller test datasets, {\it SFace} demonstrates better performance over models such as {\it VGG-Face}. {\it SFace} uses a sigmoid-constrained hypersphere loss parameter, which does leverage angular margin but in a different way than {\it ArcFace}. 

{\it GhostFaceNet} employs ``Ghost Modules,'' which are a mobile-optimized architecture and uses ArcFace (additive angular margin) for its loss function. First introduced in {\it GhostFaceNet}, Ghost Modules still perform convolution operations but at significantly less FLOPS (Floating Operations per Second). Trained on MS1MV3, it achieves competitive accuracy to other models with significantly reduced parameters and computational cost. Considering increased popularity of edge processing for FR models, {\it GhostFaceNet} is an important model to be considered in comparisons. 

The {\it buffalo\_l} and {\it antelopev2} are state-of-the-art models from the InsightFace framework \cite{insightface}. The {\it buffalo\_l} method uses ResNet-50 trained on 5.2M refined images, while {\it antelopev2} employs a Vision Transformer architecture trained on Glint360K (17M images, 360K identities) -- the largest face domain training set among evaluated models. These models represent current industry best practices and provide insight into attack transferability to modern recognition systems. Both {\it buffalo\_l} and {\it antelopev2} use ArcFace for their loss function. 

{\it DINOv3} is a state-of-the-art and the latest self-supervised, foundational, ViT-based model from Meta. The model selected for this work specifically uses \verb+dinov3-vits16-pretrain-lvd1689m+ weights (updated August 2025, \cite{dinov3_hf_model}) obtained in training the ViT on approximately 1.6 billion images from LVD-1689m dataset,
benefiting from a self-supervised (student-teacher distillation from larger model) loss.

This diversity in distance metrics and loss functions allows for evaluation of attack effectiveness across different embedding space geometries. Results are reported for each model with respect to their original loss function. These open-source models cover a decade of facial recognition advancement and provide a range of backbones, training data, and training paradigms sufficient to make general conclusions.

Diverging from the public models, {\it VeriLook} is a commercial software 
implementing proprietary algorithms and thus offering less information for evaluation. It was selected for this work since it represents a state-of-the-art commercial facial recognition technology reasonably well: (a) 
it follows ISO/IEC 30107-3 Level 2 presentation attack detection recommendations, and (b) has consistently ranked as top performer in NIST's Face Recognition Technology Evaluation (FRTE) evaluations \cite{FRTE}.

\subsection{Modeling of the IEMI Attacks}
\label{sec:modeled}

To narrow possible frequencies and modulation schemes, 
a simple attack modeling approach was developed. The modeling algorithm establishes a baseline imaging array, IEMI parameters, and video frame rate. The method computes the IEMI on array and frames per second (fps). The IEMI parameters are independently computed to overlay the image. When the sinusoid matches the same time step as the IEMI, an 8-bit discretized intensity is computed to overlay the attack effect, as shown in Fig. \ref{fig:attacked_digital}. Modeled affects were optimized for a fixed FPA of $4208\times3120$ pixels for the Samsung A26 phone (SK Hynix Hi-1339 FPA \cite{wiki:samsung_galaxy_a26}). A standard grid search of carrier frequency, amplitude, bar angle, and frequency modulation specifications was run to determine optimal disruption parameters. While certain frequencies may return equally or higher attack success, measured as increased cosine similarity between representations of clean and attack samples, other physical design constraints like hardware induction capability also drive success and were not modeled. Since proper modeling of the
full attack chain would require more complex setup of the
EMI effects on the analog circuity (specific for each FPA
and readout circuit), we decided to omit the portion in line
with traditional black box attacks, which is the focus of this paper. Therefore, the modeled parameters in the results section match the physical attack parameters: 11.465MHz carrier signal, 2Vpp (Volts peak-to-peak), and frequency modulated wave at 190kHz. 

It should also be noted the modeling is not intended to be an optimized adversarial attack. There are no machine learning models included in the modeling script. The goal is to simply generate images that have a level of qualitative disruption to provide a narrowed focus for the real-world physical attacks.  

\subsection{Data Collection}

\begin{figure}[!htbp] 
     \centering
     \begin{subfigure}[b]{0.48\linewidth} 
          \centering
          \includegraphics[width=\textwidth]{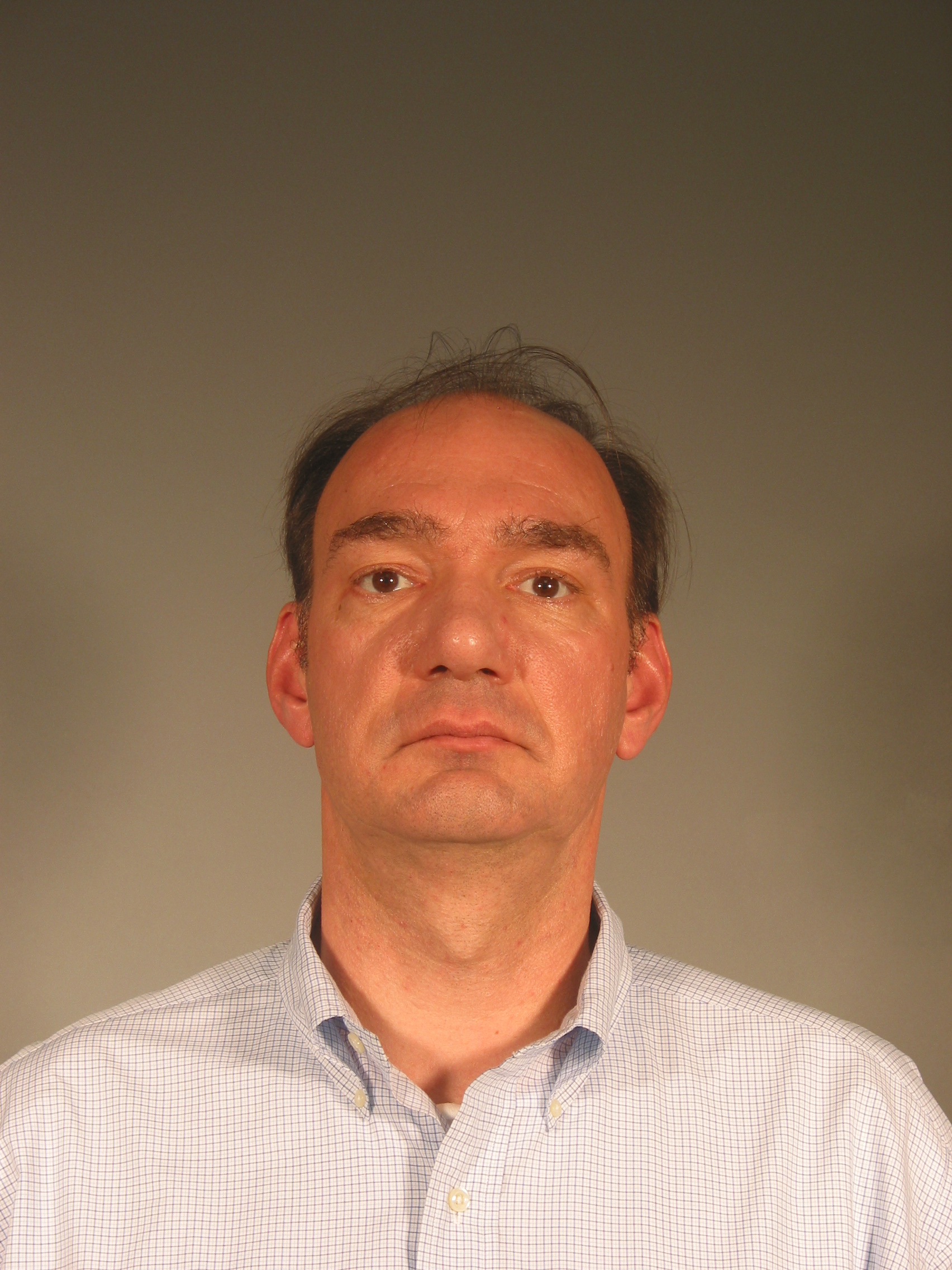}
          \caption{}
          \label{fig:original}
     \end{subfigure}
     \hfill
     \begin{subfigure}[b]{0.48\linewidth}
          \centering
          \includegraphics[width=\textwidth]{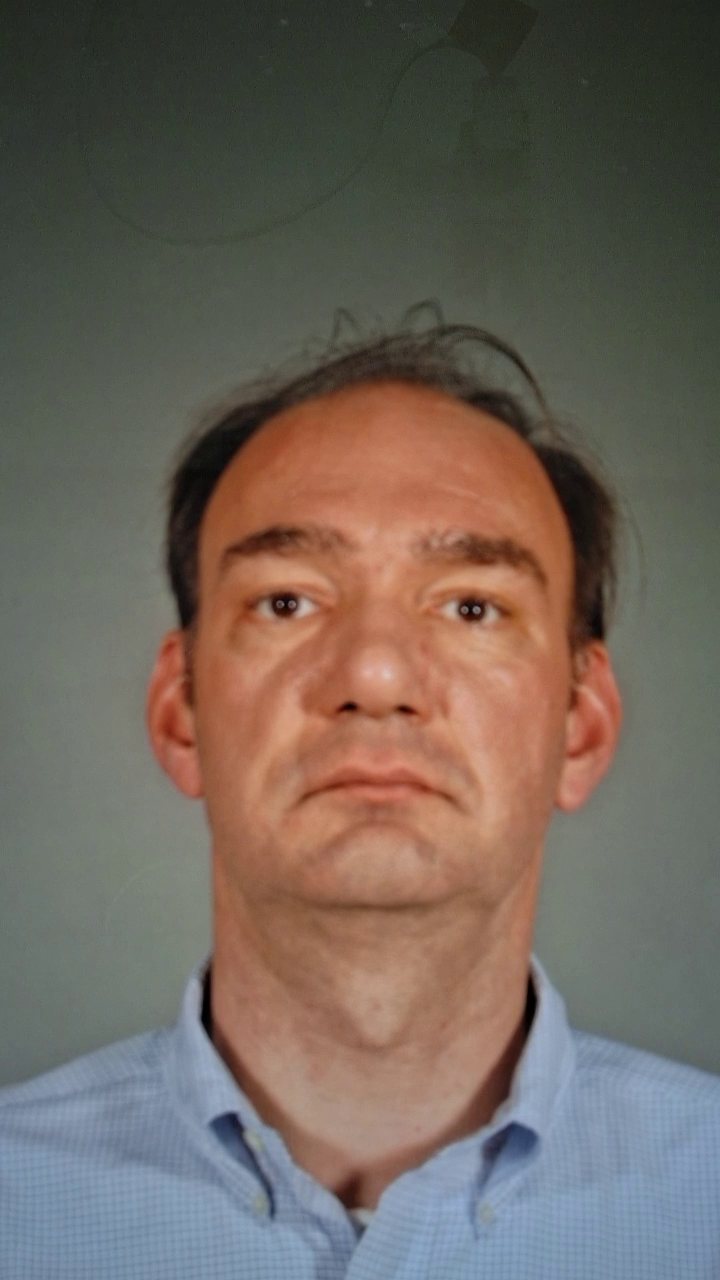}
          \caption{}
          \label{fig:clean}
     \end{subfigure}

     \vspace{5pt} 
     
     \begin{subfigure}[b]{0.48\linewidth}
          \centering
          \includegraphics[width=\textwidth]{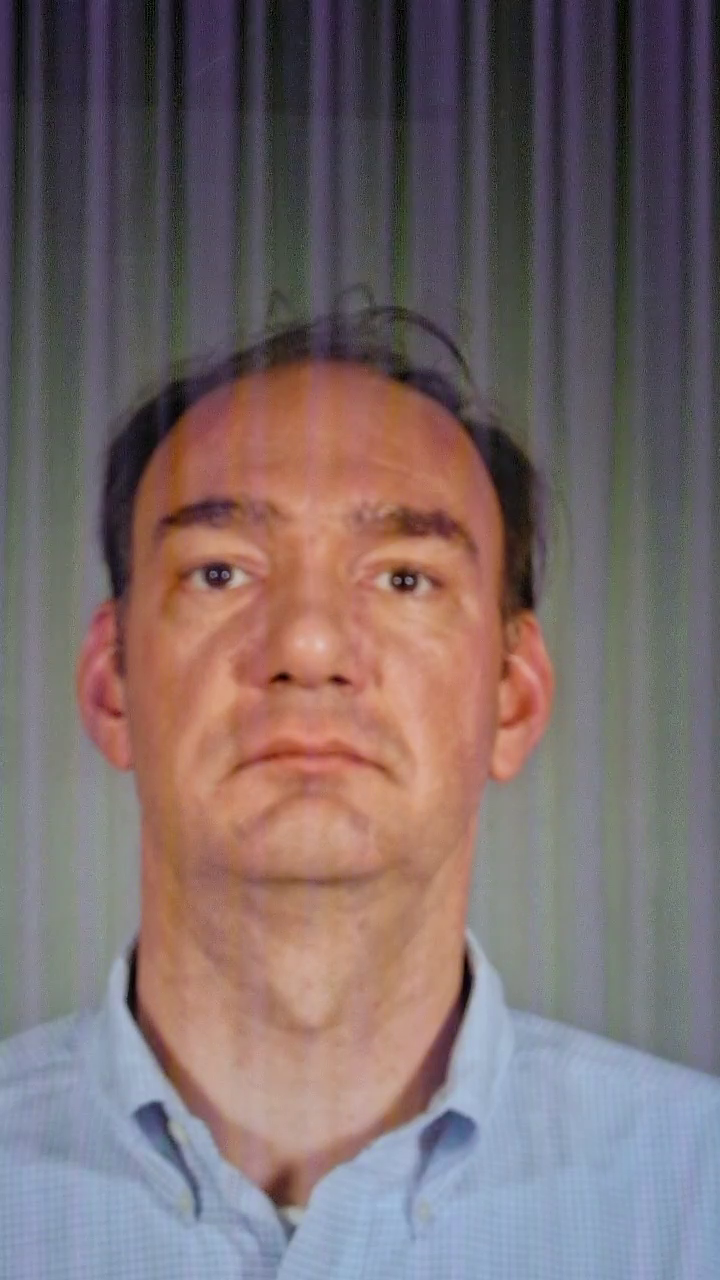}
          \caption{}
     \end{subfigure}
     \hfill
     \begin{subfigure}[b]{0.48\linewidth}
          \centering
          \includegraphics[width=\textwidth]{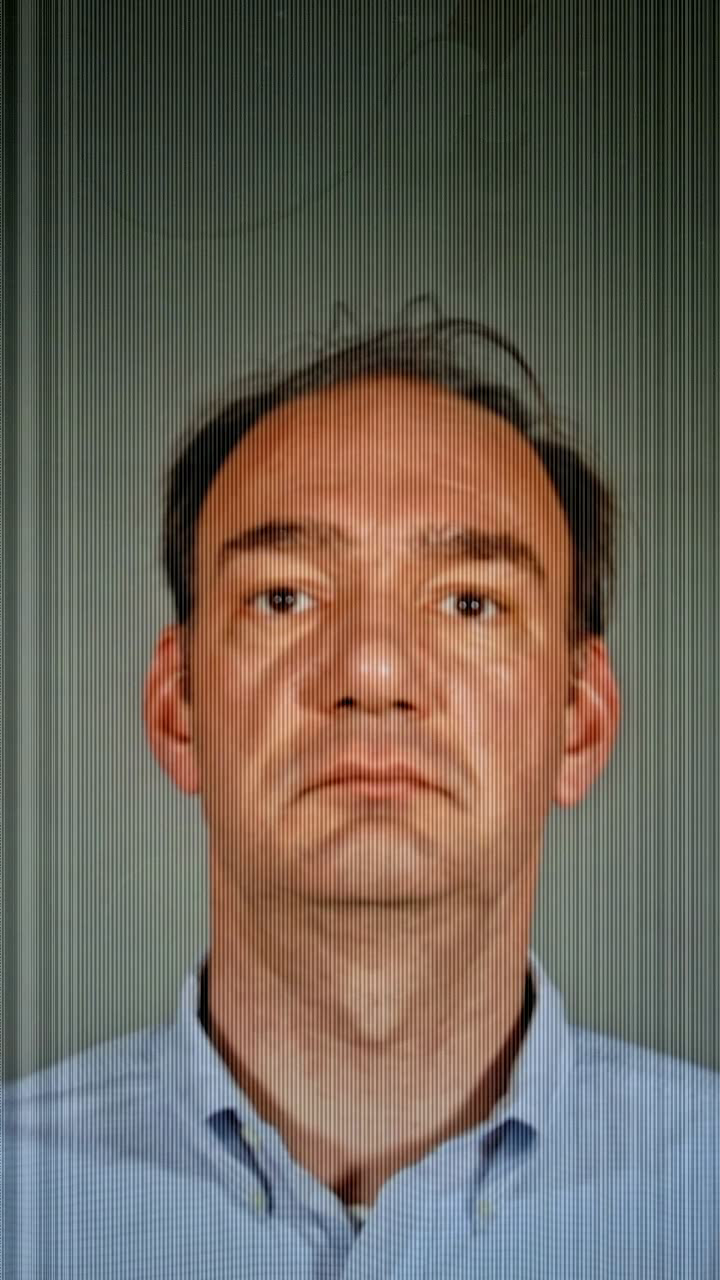}
          \caption{}
          \label{fig:attacked_digital}
     \end{subfigure}
     
     \caption{Example test images in 2x2 grid: (a) original, (b) clean, (c) physical RF attack, (d) digital RF-attack model.}
     \label{fig:four_photos}
\end{figure}

Examples of original, clean, physically attacked and modeled attack images are shown in Fig. \ref{fig:four_photos}. To collect {\bf clean samples} (Fig. \ref{fig:four_photos}b), face images representing 50 identities were displayed on the laptop screen and re-captured by the Samsung phone camera in a form of 8-second video clips (with $K=240$ frames per clip). To capture {\bf attacked images} (Fig. \ref{fig:four_photos}c), the AFG was turned on with the appropriate FM settings and connected to a copper loop. The 9 cm, single turn loop was held between 3 cm and directly on the phone, parallel to the screen and where the camera was centered in the loop. Next the same original images of the 50 identities were displayed on the laptop screen, presented to the phone at the same distance (but now with the IEMI attack active), and 8-second (= 240-frame) video clips were captured for that setup. 
Finally, the {\bf modeled attack samples} (Fig. \ref{fig:four_photos}d) were generated as described in Sec. \ref{sec:modeled} for the same set of original face images representing the same 50 identities.

We chose not to collect attack samples from human subjects, and instead imitating the clean samples as recaptures from a laptop screen. The main reason is better reproducibility of such a testing protocol, along with higher simplicity of non-human subject research. The results (see "FNMR-C" rows in Tab. \ref{table_fmr_0.1} through \ref{table_fmr_5.0}) justify this decision: testing with humans would not further prove success or failure, as the models had already low False Non-Match Rate (FNMR) values (with tight standard deviation calculated for multiple runs) on clean data captures for all baselines. In other words, while the bonafide sample is certainly a dependent variable for testing, our results show that it is less important once the models establish a reasonable FNMR.

\subsection{Physical Safety Limits}

While we did not use human subjects during experiments, we made an effort to ensure that this approach is suitable for real world deployments. The Federal Communications Committee (FCC) outlines safe exposure limits for RF testing \cite{fcc_cfr_1_1310}. Based on the measured values in Sec. \ref{sec:results}, human operators from the center of the copper loop should be no closer than 1.6 feet for occupational works and 2.1 feet away for the general population. These distances are also time-dependent averages, where occupational workers are averaged over six minutes and general population over 30 minutes. Based on our experimental setup and attack time scale, we determine this to still be an acceptable setup for real world attack applications, especially considering additional mitigation that can be implemented (RF shielding, different antenna design, remote control of attack hardware). In addition to human safety distances, the laptop used during experimentation was outside 1.6 feet radius and had no adverse IEMI affects.  

\subsection{Performance Metrics}

Since we do not assess the presentation attack detection in this work, and rather we evaluate the matching performance deterioration under the RF attacks, we are not using ISO/IEC 30107 PAD-specific metrics such as APCER and BPCER \cite{ISO30107-3:2023}. Instead, for evaluation of the attack success, we report False Non-Match Rate (FNMR) at several defined False Match Rate (FMR) values: 0.1\%, 1\% and 5\%, and compare the FNMR values across two scenarios (when face recognition methods are fed with clean images and physical attack images). Note an exception for VeriLook, as the commercial software does not go above FMR=0.1\% (software-defined hard limit), therefore it is only evaluated at a single threshold. 

First, to set the appropriate acceptance thresholds associated with FMR=\{0.1\%,1\%,5\%\}, we generated $N(N-1)/2 = 1,225$ impostor comparisons per model (using either Euclidean distance or cosine similarity, as appropriate) using original images from the MBGCv2 dataset representing $N=50$ unique identities used in this study.  

Next, for each 8-second video clip we calculate subject-level FNMR$_n$, by generating all possible genuine scores for that subject $n$ (except for symmetric comparisons, thus calculating $K(K-1)/2$ genuine comparison scores, where $K$ is the number of available same-clip frames), and applying acceptance thresholds appropriate for the assumed FMR levels.

Then, we define attack success for subject $n$ if $\text{FNMR}_{n,\text{clean}} < \text{FNMR}_{n,\text{attack}}$ holds true. We calculate the overall Physical Attack Success (PAS) as the rate of the number of subjects for whom the attack was successful, namely:

$$
\text{PAS} = \frac{\big|\{n : \text{FNMR}_{n,\text{attack}} > \text{FNMR}_{n,\text{clean}}\}\big|}{N},
$$

\noindent
where $n=1,\dots,N$, and $|\cdot|$ denotes the number of elements in a set. In our calculations N=50.

To assess the statistical significance of the differences between PAS point estimators, we randomly sampled 120 frames 10 times from the 240 available frames and provide both mean and standard deviation values over these 10 runs.

The overall FNMR$_\text{clean}$ and FNMR$_\text{attack}$ are computed using 50\% (\ie, 120) randomly picked frames from all experiment 8-second clips (\ie, 50 identities $\times$ 240 frames = 12,000 data points in total). Again, to assess the statistical significance of the differences between FNMR values, we repeated this sampling (with replacement) 10 times, and report mean and standard deviation of the FNMR values.

Finally, the ESR (Evasion Success Rate) computes how frequently an attacked identity is accepted as a new identity, under the appropriate FMR level.

\section{Results and Discussion}
\label{sec:results}

\begin{table}[!t]
    \centering
    \renewcommand{\arraystretch}{0.85} 
    \setlength{\tabcolsep}{3pt}
    
    \begin{threeparttable}
        \footnotesize 
        \caption{Results (in \%) for the acceptance threshold at FMR=0.1\%.}
        \label{table_fmr_0.1}
        \begin{tabular}{l *{7}{c}}
            \toprule
            Metric & \rotatebox{60}{VGG-Face} & \rotatebox{60}{SFace} & \rotatebox{60}{\makecell{Ghost\\FaceNet}} & \rotatebox{60}{buffalo\_l} & \rotatebox{60}{antelopev2} & \rotatebox{60}{VeriLook} & \rotatebox{60}{DINOv3} \\
            \midrule
            \makecell[l]{FNMR$_\text{clean}$ \\ ($\pm$ std)} & \makecell{5.2 \\ $\pm$2.7} & \makecell{6.1 \\ $\pm$2.8} & \makecell{6.0 \\ $\pm$3.5} & \makecell{0.0 \\ $\pm$0} & \makecell{0.0 \\ $\pm$0.0} & \makecell{0.0 \\ $\pm$0.0} & \makecell{0.66 \\ $\pm$4.3}\\
            \addlinespace[2pt]
            \makecell[l]{FNMR$_\text{attack}$ \\ ($\pm$ std)} & \makecell{7.58 \\ $\pm$19.8} & \makecell{8.12 \\ $\pm$20.1} & \makecell{8.4 \\ $\pm$19.9} & \makecell{0.0 \\ $\pm$0.0} & \makecell{0.0 \\ $\pm$0.0} & \makecell{22.0 \\ $\pm$0.0} & \makecell{\textbf{100.0} \\ $\pm$0.0}\\
            \addlinespace[2pt]
            \midrule
            \makecell[l]{PAS \\ ($\pm$ std)} & \makecell{56.0 \\ $\pm$14.9} & \makecell{56.0 \\ $\pm$14.9} & \makecell{58.0 \\ $\pm$15.4} & \makecell{0.0 \\ $\pm$0.0} & \makecell{0.0 \\ $\pm$0.0} & \makecell{22.0 \\ $\pm$0.0} & \makecell{\textbf{96.0} \\ $\pm$0.0}\\
            \makecell[l]{ESR} & \makecell{\textbf{14.0}} & \makecell{0.0} & \makecell{0.0} & \makecell{0.0} & \makecell{0.0} & \makecell{2.0} & \makecell{0.0}\\
            \addlinespace[2pt]
            \bottomrule
        \end{tabular}
    \end{threeparttable}

    \vspace{1.2em}

    \begin{threeparttable}
        \footnotesize 
        \caption{Same as in Tab. \ref{table_fmr_0.1}, except that FMR=1\%}
        \label{table_fmr_1.0}
        \begin{tabular}{l *{6}{c}}
            \toprule
            Metric & \rotatebox{60}{VGG-Face} & \rotatebox{60}{SFace} & \rotatebox{60}{\makecell{Ghost\\FaceNet}} & \rotatebox{60}{buffalo\_l} & \rotatebox{60}{antelopev2} & \rotatebox{60}{DINOv3}\\
            \midrule
            \makecell[l]{FNMR$_\text{clean}$ \\ ($\pm$ std)} & \makecell{7.2 \\ $\pm$2.1} & \makecell{7.2 \\ $\pm$2.5} & \makecell{7.6 \\ $\pm$2.5} & \makecell{0.0 \\ $\pm$0.0} & \makecell{0.0 \\ $\pm$0.0} & \makecell{0.06 \\ $\pm$0.31} \\
            \addlinespace[2pt]
            \makecell[l]{FNMR$_\text{attack}$ \\ ($\pm$ std)} & \makecell{7.3 \\ $\pm$19.6} & \makecell{7.9 \\ $\pm$17.8} & \makecell{7.1 \\ $\pm$17.2} & \makecell{0.0 \\ $\pm$0.0} & \makecell{0.0 \\ $\pm$0.0} & \makecell{\textbf{98.3} \\ $\pm$10.7}\\
            \addlinespace[2pt]
            \midrule
            \makecell[l]{PAS \\ ($\pm$ std)} & \makecell{52.0 \\ $\pm$14.6} & \makecell{54.0 \\ $\pm$14.9} & \makecell{54.0 \\ $\pm$14.9} & \makecell{0.0 \\ $\pm$0.0} & \makecell{0.0 \\ $\pm$0.0} & \makecell{\textbf{86.0} \\ $\pm$0.0}\\
            \makecell[l]{ESR} & \makecell{30.0} & \makecell{\textbf{54.0}} & \makecell{\textbf{54.0}} & \makecell{0.0} & \makecell{0.0} & \makecell{0.0}\\
            \addlinespace[2pt]
            \bottomrule
        \end{tabular}
    \end{threeparttable}

    \vspace{1.2em}

    \begin{threeparttable}
        \footnotesize 
        \caption{Same as in Tab. \ref{table_fmr_0.1}, except that FMR=5\%}
        \label{table_fmr_5.0}
        \begin{tabular}{l *{6}{c}} 
            \toprule
            Metric & \rotatebox{60}{VGG-Face} & \rotatebox{60}{SFace} & \rotatebox{60}{\makecell{Ghost\\FaceNet}} & \rotatebox{60}{buffalo\_l} & \rotatebox{60}{antelopev2} & \rotatebox{60}{DINOv3} \\
            \midrule
            \makecell[l]{FNMR$_\text{clean}$ \\ ($\pm$ std)} & \makecell{3.9 \\ $\pm$1.5} & \makecell{6.1 \\ $\pm$3.1} & \makecell{6.2 \\ $\pm$1.8} & \makecell{0.0 \\ $\pm$0.0} & \makecell{0.0 \\ $\pm$0.0} & \makecell{0.03 \\ $\pm$0.19}\\
            \addlinespace[2pt]
            \makecell[l]{FNMR$_\text{attack}$ \\ ($\pm$ std)} & \makecell{6.4 \\ $\pm$16.4} & \makecell{7.3 \\ $\pm$17.6} & \makecell{7.1 \\ $\pm$17.1} & \makecell{0.0 \\ $\pm$0.0} & \makecell{0.0 \\ $\pm$0.0} & \makecell{\textbf{73.2 }\\ $\pm$37.2}\\
            \midrule
            \makecell[l]{PAS \\ ($\pm$ std)} & \makecell{52.0 \\ $\pm$14.6} & \makecell{\textbf{54.0} \\ $\pm$14.8} & \makecell{\textbf{54.0} \\ $\pm$14.8} & \makecell{0.0 \\ $\pm$0.0} & \makecell{0.0 \\ $\pm$0.0} & \makecell{40.0 \\ $\pm$0.0}\\ 
            \makecell[l]{ESR} & \makecell{\textbf{96.0}} & \makecell{88.0} & \makecell{ 88.0} & \makecell{0.0} & \makecell{0.0} & \makecell{0.0}\\
            \addlinespace[2pt]
            \bottomrule
        \end{tabular}
    \end{threeparttable}
\end{table}

\subsection{Are the FM attacks successful? And if so, why?}
The performance results are presented in Tables \ref{table_fmr_0.1}-\ref{table_fmr_5.0}, from which we can conclude that using IEMI to attack facial recognition systems is an effective method. 

The presented approach had varying levels of success across FMR values and models. While multiple frequencies and FM settings resulted in successful perturbations, the settings reported in this paper point to \textit{what} in the optical and processing train was attacked. The front facing camera on the Samsung A26 has a sampled resolution of $1920\times1080$ pixels ($4208\times3120$ pixels on the FPA) and records at 30 fps. The general starting point for center point frequencies that could cause disruption then is 62.208 MHz, determined by multiplying all $1920*1080*30$ together. Since imaging integration time can affect readout speeds, the frequency is a starting point and our reported carrier frequency of 11.465 MHz is not a direct harmonic. The MHz interference impacts the pixel clock timing, which translates to the vertical bars seen in Fig. \ref{fig:four_photos}c. The vertical lines appear to adversely affect the early-layers convolutional filters. 

The modulation adds another aspect. Utilizing a frequency modulated triangle wave at 190 kHz, the vertical bars now sweep and vibrate across the image. This aliasing effect is likely due to disrupted timing of the Horizontal Timing Control (HTC) within the readout circuit. The 11.465 MHz affect can be isolated without modulation to see a static set of vertical bars. Conversely, the FM can be isolated by changing the carrier frequency to a minimal interference and still observing the HTC disruption.

To understand why the attack works at a fundamental level, Maxwell's Law is the most appropriate explainability tool. According to Maxwell's Law, the time-varying current injected into the copper loop generates both electric and magnetic fields. In the near-field region of our loop, the magnetic component is expected to be the primary vector for induction. During testing the loop orientation was changed between parallel and perpendicular with respect to the phone but the reported results focus solely on parallel experiments. Results are focused on parallel experiments since the electromagnetic field is strongest when the coil is parallel to the phone. Measuring the center of the loop provided a magnetic field of approximately 175 mGauss (mG) $\pm$15 mG and an electric field strength of 192 Volts/meter (V/m) $\pm$15 V/m. For comparison, a generic household hairdryer can produce 30-50 V/m of electric field and between 100-700 mG of magnetic field, but our device produces these values at higher, critical interference frequencies \cite{GreenFactsMagneticFields}. These values and orientations are important to understand which field is producing the dominant effect for reproducibility and transferability.

\subsection{Generalization Across Methods}

The results also suggest that the attack generalizes across backbones and loss functions, with the exception of ArcFace or angular loss. As evident by the models utilizing ArcFace specifically, they are completely resilient to the attack. Conversely, a state-of-the-art ViT-based foundational model (included into DINOv3 suite), was the \textit{most} susceptible to the attack in terms of FNMR. We posit the loss function specificity is due to the attack structure, where ArcFace is most discriminatory against subtle changes in identities. Models trained with other methods such as cross-entropy loss, triplet loss, and self-supervised (with student-teacher distillation) offer a mix of older approaches and new approaches meant for generalization. The partial exception to this observation is {\it GhostFaceNet}, which does use ArcFace loss. The major different between {\it GhostFaceNet} and the other ArcFace loss models is their respective learning capacity. {\it GhostFaceNet} is meant for edge processing and only possess 0.82M parameters, whereas {\it buffalo\_l} has 25M and {\it antelopev2} has 65M parameters. Therefore, while the loss function is working hard to repel the attack in {\it GhostFaceNet}, the tradeoff in increased performance appears to be attack resistance. While we do not have the same level of inspection for {\it VeriLook} compared to the other open source models, it was still vulnerable to the attack. Interestingly, when {\it VeriLook} was unable to correctly match an identity for a given attack video, the model failed for all frames in the video sequence, which is why FNMR$_\text{attack}$ and PAS match. Matching FNMR$_\text{attack}$ and PAS was unique to {\it VeriLook}. 

The ESR (Evasion Success Rate) was quite successful across {\it VGG-Face}, {\it SFace}, and {\it GhostFaceNet}. {\it VGG-Face} was the only model that showed non-zero ESR values across all FMR levels and had an incredible 96\% success rate at FMR=5\%. A visual of {\it GhostFaceNet} embeddings is shown in Fig. \ref{fig:AttackTrajectories}-\ref{fig:UntargetedAttack}. The blue circles (clean images) and red X's (attacked images) are often shown close together but further inspection shows the attacks move one identity close to another for ESR.

\subsection{Generalization Across Hardware}
Existing literature has shown physical attacks generalize across similar cameras or sensors (\eg, resolution, readout frequency, manufacturer \cite{Kohler2021TheySeeMe,liu2025}). Due to testing precedent, we hypothesize the attack generalizes to other smartphones or cameras using the Samsung Galaxy A26's 13MP Hynix family CMOS. The Hynix CMOS is found in multiple Samsung Galaxy series smartphones \cite{GSMArena2025Samsung}.

\subsection{Frequency Analysis}

\begin{figure}[!htbp] 
     \centering
     \begin{subfigure}[b]{0.31\linewidth}
          \centering
          \includegraphics[width=\textwidth]{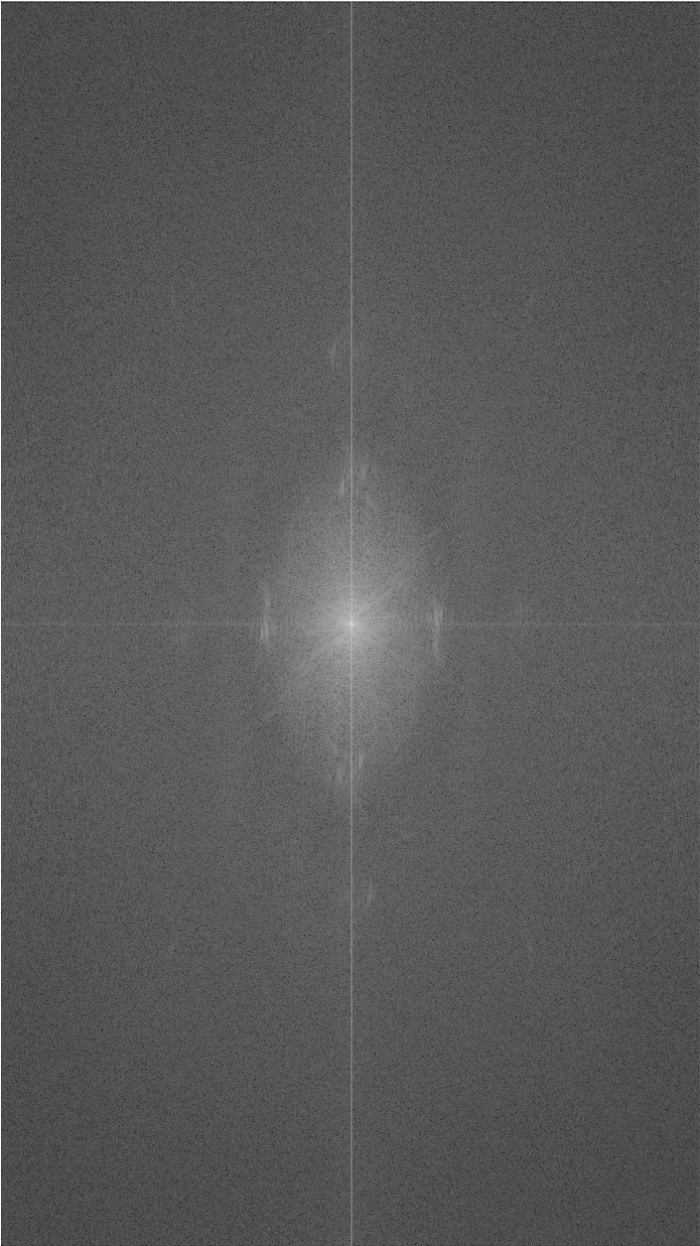}
          \caption{}
          \label{fig:FFToriginal}
     \end{subfigure}
     \hfill
     \begin{subfigure}[b]{0.31\linewidth}
          \centering
          \includegraphics[width=\textwidth]{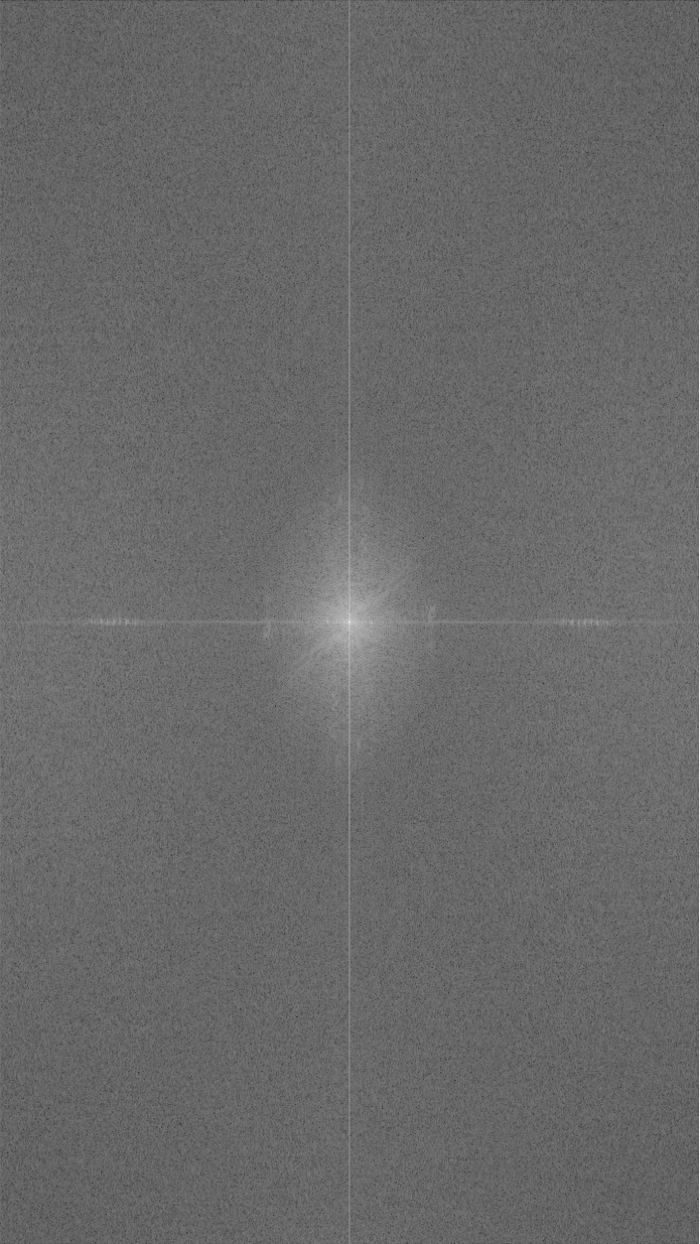}
          \caption{}
          \label{fig:FFTphysical}
     \end{subfigure}
     \hfill
     \begin{subfigure}[b]{0.31\linewidth}
          \centering
          \includegraphics[width=\textwidth]{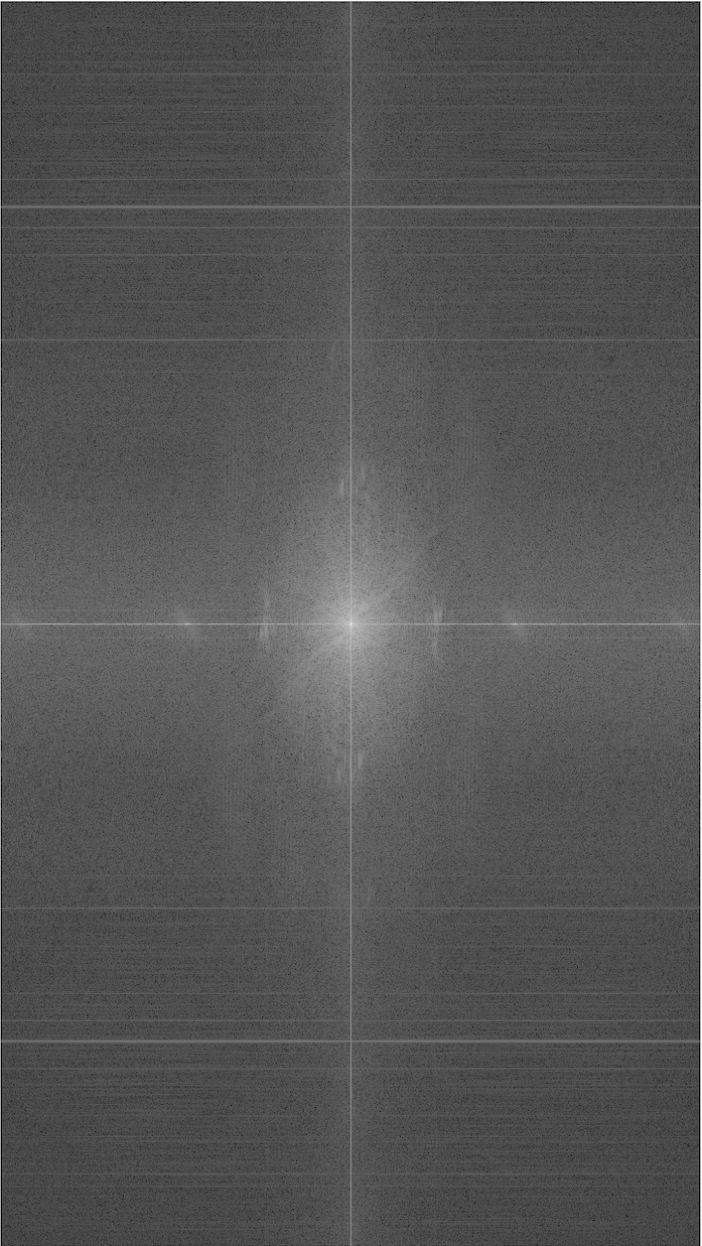}
          \caption{}
          \label{fig:FFTmodeled}
     \end{subfigure}
     
     \caption{FFT amplitude spectra calculated for the same identity as in Fig.~\ref{fig:four_photos}: (a) clean MBGCv2 sample, (b) physically attacked sample, and (c) modeled attack (as described in Sec.~\ref{sec:modeled}).}
     \label{fig:FFT_group}
\end{figure}

In addition to examining the FNMR and PAS, we analyzed the attacks via 2D-FFT (Fast Fourier Transform). The images shown in Fig. \ref{fig:FFToriginal}-\ref{fig:FFTphysical} visually demonstrate how the modeled images are \textit{ideal} attacks and create uniform frequencies over the existing clean image. These results demonstrate where the modeled effect can be global in the image. Fig. \ref{fig:FFTphysical} shows a rather smooth frequency amplitude spectrum, though the attack has increased the overall image intensity.

\subsection{Model Embedding Space Analysis}

\begin{figure}[!htbp] 
     \centering
     \begin{subfigure}[b]{1.0\linewidth}
          \centering
          \includegraphics[width=\textwidth]{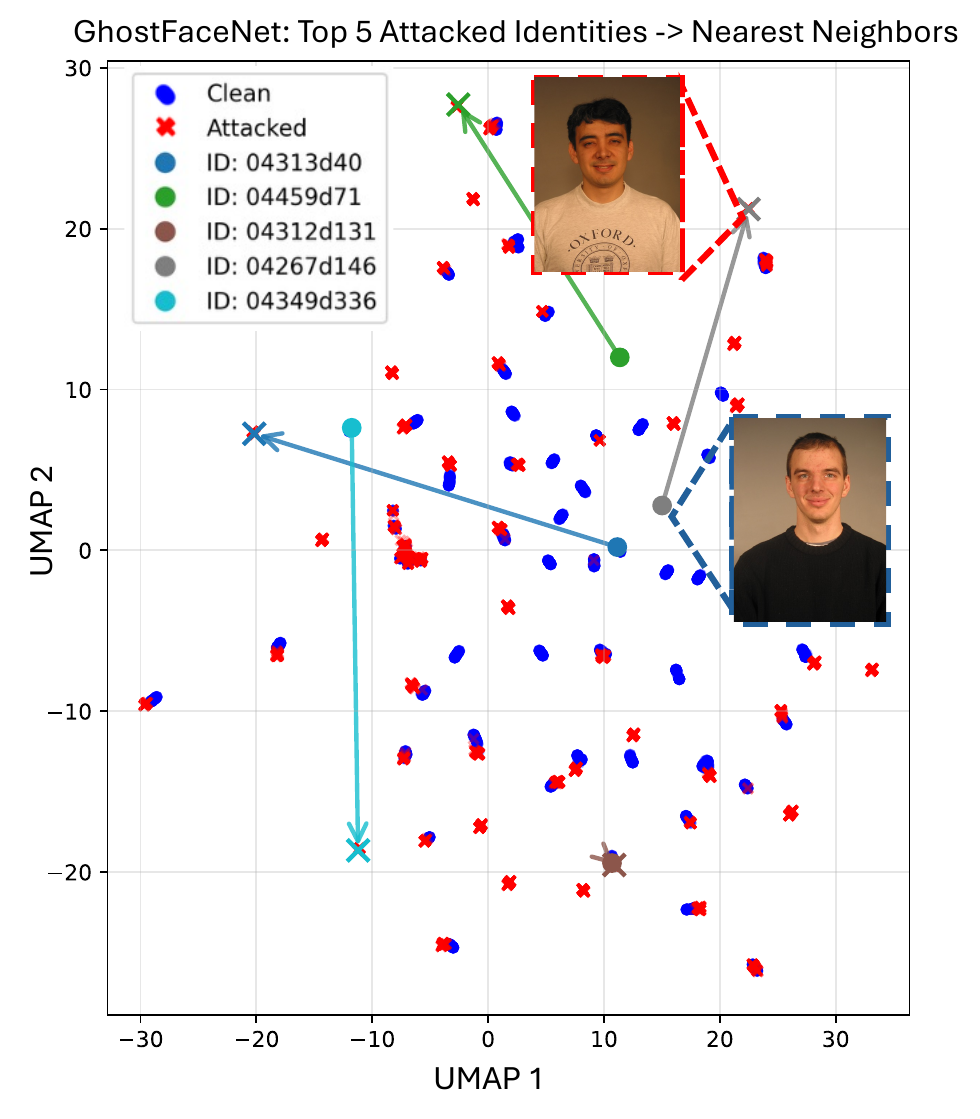}
          \caption{}
          \label{fig:AttackTrajectories}
     \end{subfigure}
     
     \vspace{1em} 

     \begin{subfigure}[b]{1.1\linewidth}
          \centering
          \includegraphics[width=\textwidth]{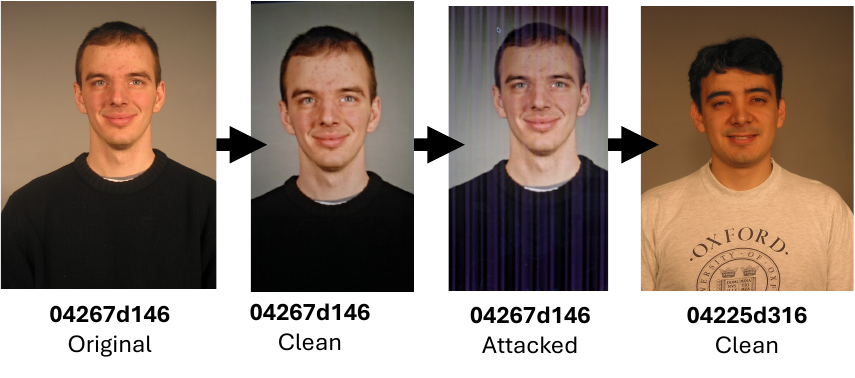}
          \caption{}
          \label{fig:UntargetedAttack}
     \end{subfigure}
     
     \caption{(a) UMAP projection of the representations of top 5 attacked identities within the {\it GhostFaceNet} embedding space (at FMR=5.0\%), and (b) the example illustration of successful untargeted attack. Remaining plots for all models are provided in Supplementary Materials.}
     \label{fig:combined_vertical}
\end{figure}

Another way to visualize the success the of the attack is through UMAP (Uniform Manifold Approximation and Projection) plots, using {\it GhostFaceNet} at FMR=5.0\%. Fig. \ref{fig:AttackTrajectories} shows the five identities with the largest differences in euclidean distance or cosine similarity, measured between the clean and attacked data. The purple dots are clean image identities and the attacked identities are orange dots. The UMAP tracks the embedding trajectory of the five identities with the largest attack success. To further illustrate why this untargeted attack can be quite successful, Fig. \ref{fig:AttackTrajectories} shows those same five identities and their \textit{new} nearest neighbors within the clean image dataset. 

As it can be seen, all attacked images have ``migrated'' from their original identity to new identities \textit{and} have been accepted as different identities by satisfying the {\it GhostFaceNet} FMR-level-defined acceptance threshold. This means the attack not only changed the original identity but, in an untargeted attack, pushed the image close enough to a different identity to then be successfully classified. This untargeted attack success is a subset of the PAS in Tables \ref{table_fmr_0.1}-\ref{table_fmr_5.0}. We additionally visualize this affect in Fig. \ref{fig:UntargetedAttack}, where identity \#04267d146 is attacked and then successfully becomes identity \#04225d316 in the MBGCv2 dataset. 

\section{Conclusions}
This work presents, to our knowledge, the first intentional electromagnetic interference (IEMI) attack on machine learning-based face recognition systems. Further, this work examines the effects of IEMI on CMOS detectors. The IEMI attack was successful in a black box setting across five models (out of seven methods used in evaluations). The attacks success across multiple backbones and loss functions, demonstrating a clear vulnerability in existing methods. The best defense is to use a model with ArcFace as its loss function. While we tested a commercial phone without any modifications, further investigation is required to determine breadth of hardware vulnerability. 

We offer source codes of the attack modeling technique, as well as clean and under-attack face videos with this paper 
to facilitate reproducibility and future research in this area.

Finally, the results presented in this paper may stimulate adding the RF-based attacks to the portfolio of presentation attacks in developments of future versions of recommendation, such as ISO/IEC 30107 \cite{ISO30107-3:2023} or FIDO Alliance Face Verification Certification Program \cite{fido}. 

\section{Acknowledgments}

This material is based upon work partially supported by the OUSW/R\&E (Office of the Under Secretary of War, Research and Engineering), National Defense Education Program (NDEP) SMART Scholarship Program, and Naval Surface Warfare Center (NSWC), Crane Division Ph.D. Fellowship Program. Any opinions, findings, and conclusions or recommendations expressed in this material are those of the authors and do not necessarily reflect the views of the DoW or U.S. Navy.

{\small
\bibliographystyle{ieee}
\bibliography{egbib}

\begin{thebibliography}{10}\itemsep=-1pt

\bibitem{Akhtar2018Survey}
N.~Akhtar and A.~Mian.
\newblock Threat of adversarial attacks on deep learning in computer vision: A survey.
\newblock {\em IEEE Access}, 6:14410--14430, 2018.

\bibitem{Alansarietal2023}
M.~Alansari, O.~Abdul~Hay, S.~Javed, A.~Shoufan, Y.~Zweiri, and N.~Werghi.
\newblock Ghostfacenets: Lightweight face recognition model from cheap operations.
\newblock {\em IEEE Access}, 11:43447--43461, 2023.

\bibitem{rf_amp_1_930mhz}
{Amazon.com}.
\newblock 1-930{MHz} 2.0{W} professional {RF} power amplifier module.
\newblock \url{https://www.amazon.com/dp/B09HX3C43K}, 2026.
\newblock Accessed: 2026-04-18.

\bibitem{Antil_Neurocomputing_2025}
A.~Antil and C.~Dhiman.
\newblock Unmasking deception: A comprehensive survey on the evolution of face anti‐spoofing methods.
\newblock {\em Neurocomputing}, 617:128992, 2025.

\bibitem{bochkovskiy2020yolov4}
A.~Bochkovskiy, C.-Y. Wang, and H.-Y.~M. Liao.
\newblock Yolov4: Optimal speed and accuracy of object detection.
\newblock {\em arXiv preprint arXiv:2004.10934}, 2020.

\bibitem{Caoetal2018}
Q.~Cao, L.~Shen, W.~Xie, O.~M. Parkhi, and A.~Zisserman.
\newblock Vggface2: A dataset for recognising faces across pose and age.
\newblock In {\em 2018 13th IEEE International Conference on Automatic Face \& Gesture Recognition (FG 2018)}, 2018.

\bibitem{fcc_cfr_1_1310}
{Federal Communications Commission}.
\newblock 47 {CFR} § 1.1310 - {R}adiofrequency radiation exposure limits.
\newblock Code of Federal Regulations, Title 47, Volume 1, 2011.
\newblock Accessed: April 17, 2026.

\bibitem{fido}
{FIDO Alliance}.
\newblock Face verification certification.
\newblock \url{https://fidoalliance.org/certification/identity-verification/face-verification/}, 2026.

\bibitem{goiffon2023}
V.~Goiffon.
\newblock Radiation effects on cmos active pixel image sensors.
\newblock Presented at IEEE Nucl. Space Radiat. Effects Conf., 2023.

\bibitem{GreenFactsMagneticFields}
{GreenFacts}.
\newblock Magnetic fields generated by domestic appliances, 2004.
\newblock Based on the IARC (2002) and California EMF Program (2002) reports.

\bibitem{GSMArena2025Samsung}
{GSMArena Team}.
\newblock Samsung {G}alaxy {A}26 review: Camera.
\newblock GSMArena, Mar. 2025.
\newblock Accessed: 2026-07-09.

\bibitem{insightface}
J.~Guo, J.~Deng, A.~Lattas, and S.~Zafeiriou.
\newblock Insightface: 2d and 3d face analysis project.
\newblock \url{https://github.com/deepinsight/insightface}, 2021.

\bibitem{he2016deep}
K.~He, X.~Zhang, S.~Ren, and J.~Sun.
\newblock Deep residual learning for image recognition.
\newblock In {\em Proceedings of the IEEE Conference on Computer Vision and Pattern Recognition (CVPR)}, pages 770--778, 2016.

\bibitem{ISO30107-3:2023}
Information technology -- biometric presentation attack detection -- part 3: Testing and reporting.
\newblock Standard ISO/IEC 30107-3:2023, International Organization for Standardization, Geneva, CH, 2023.

\bibitem{ji2021poltergeist}
X.~Ji, Y.~Cheng, Y.~Cheng, K.~Wang, and W.~Xu.
\newblock Poltergeist: Acoustic adversarial machine learning against cameras and computer vision.
\newblock In {\em 2021 IEEE Symposium on Security and Privacy (SP)}, pages 101--118. IEEE, 2021.

\bibitem{Kohler2021TheySeeMe}
S.~K{\"{o}}hler, G.~Lovisotto, S.~Birnbach, R.~Baker, and I.~Martinovic.
\newblock They see me rollin': Inherent vulnerability of the rolling shutter in cmos image sensors.
\newblock In {\em Proceedings of the 37th Annual Computer Security Applications Conference}, ACSAC '21, pages 399--413, New York, NY, USA, 2021. Association for Computing Machinery.

\bibitem{leyva2023}
R.~Leyva.
\newblock Attacks against face recognition systems: A state-of-the-art review.
\newblock Technical report, The Alan Turing Institute, 2023.

\bibitem{liu2025}
Z.~Liu, F.~Lin, Z.~Ba, L.~Lu, and K.~Ren.
\newblock Magshadow: Physical adversarial example attacks via electromagnetic injection.
\newblock {\em IEEE Transactions on Dependable and Secure Computing}, 22(4):3307, -07 2025.

\bibitem{dinov3_2025}
{Meta AI}.
\newblock {DINOv3}: Foundation models producing excellent dense features.
\newblock {\em arXiv preprint arXiv:2508.10104}, 2025.

\bibitem{dinov3_hf_model}
{Meta AI}.
\newblock Dinov3 {ViT-L/16} pre-trained on {LVD-1689M}.
\newblock \url{https://huggingface.co/facebook/dinov3-vitl16-pretrain-lvd1689m}, 2025.
\newblock Hugging Face Model Hub.

\bibitem{verilook}
{Neurotechnology}.
\newblock {\em VeriLook SDK: Face Identification Technology}.
\newblock Neurotechnology, Vilnius, Lithuania, 2024.
\newblock Available at \url{https://www.neurotechnology.com/verilook.html}.

\bibitem{Nguyenetal2020}
D.-L. Nguyen, S.~S. Arora, Y.~Wu, and H.~Yang.
\newblock Adversarial light projection attacks on face recognition systems: A feasibility study.
\newblock In {\em 2020 IEEE/CVF Conference on Computer Vision and Pattern Recognition Workshops (CVPRW)}, pages 3548--3556. IEEE, 2020.

\bibitem{FRTE}
{NIST}.
\newblock {Face Technology Evaluations -- FRTE/FATE}.
\newblock \url{https://www.nist.gov/programs-projects/face-technology-evaluations-frtefate }, 2026.
\newblock Accessed: 2026-04-27.

\bibitem{Ramachandra_CSUR_2017}
R.~Ramachandra and C.~Busch.
\newblock Presentation attack detection methods for face recognition systems: A comprehensive survey.
\newblock {\em ACM Computing Surveys}, 50(1):1–37, Mar. 2017.

\bibitem{redmon2018yolov3}
J.~Redmon and A.~Farhadi.
\newblock {YOLO v3: An incremental improvement}.
\newblock {\em arXiv preprint arXiv:1804.02767}, 2018.

\bibitem{ren2025ghostshot}
Y.~Ren, Q.~Jiang, C.~Yan, X.~Ji, and W.~Xu.
\newblock Ghostshot: Manipulating the image of ccd cameras with electromagnetic interference.
\newblock In {\em Proceedings of the 32nd Network and Distributed System Security (NDSS) Symposium}, San Diego, CA, USA, February 2025. Internet Society.

\bibitem{samsung_galaxy_a26}
{Samsung Electronics}.
\newblock Galaxy {A26} 5g.
\newblock \url{https://www.samsung.com/us/smartphones/galaxy-a26-5g/}, 2025.
\newblock Accessed: 2026-04-18.

\bibitem{sayles2021invisible}
A.~Sayles, A.~Hooda, M.~Gupta, R.~Chatterjee, and E.~Fernandes.
\newblock Invisible perturbations: Physical adversarial examples exploiting the rolling shutter effect.
\newblock In {\em Proceedings of the IEEE/CVF Conference on Computer Vision and Pattern Recognition (CVPR)}, pages 14666--14675, 2021.

\bibitem{sharif2016}
M.~Sharif, S.~Bhagavatula, L.~Bauer, and M.~K. Reiter.
\newblock Accessorize to a crime.
\newblock In {\em Proceedings of the 2016 ACM SIGSAC Conference on Computer and Communications Security}, page 1528, 2025-10-24 2016.

\bibitem{szegedy2016rethinking}
C.~Szegedy, V.~Vanhoucke, S.~Ioffe, J.~Shlens, and Z.~Wojna.
\newblock Rethinking the inception architecture for computer vision.
\newblock In {\em Proceedings of the IEEE Conference on Computer Vision and Pattern Recognition (CVPR)}, pages 2818--2826, 2016.

\bibitem{tek_afg31000}
{Tektronix, Inc.}
\newblock Afg31000 series arbitrary function generator.
\newblock \url{https://www.tek.com/en/products/signal-generators/arbitrary-function-generator/afg31000}, 2026.
\newblock Accessed: 2026-04-18.

\bibitem{wiki:samsung_galaxy_a26}
{Wikipedia contributors}.
\newblock {Samsung Galaxy A26 5G}, 2026.

\bibitem{Yu_TPAMI_2022}
Z.~Yu, Y.~Qin, X.~Li, C.~Zhao, Z.~Lei, and G.~Zhao.
\newblock Deep learning for face anti-spoofing: A survey.
\newblock {\em IEEE Transactions on Pattern Analysis and Machine Intelligence}, page 1–22, 2022.

\bibitem{zhang2025rainbow}
Y.~Zhang, L.~Wang, S.~Chen, and J.~Liu.
\newblock Rainbow artifacts from electromagnetic signal injection attacks on image sensors.
\newblock {\em arXiv preprint arXiv:2507.07773}, 2025.

\bibitem{Zhongetal2021}
Y.~Zhong, W.~Deng, J.~Hu, D.~Zhao, X.-S. Li, and H.~Wen.
\newblock Sface: Sigmoid-constrained hypersphere loss for robust face recognition.
\newblock {\em IEEE Transactions on Image Processing}, 30:2587--2598, 2021.

\end{thebibliography}
}

\clearpage
\onecolumn

\setcounter{section}{0}
\setcounter{figure}{0}
\setcounter{table}{0}
\setcounter{equation}{0}

\thispagestyle{empty} 

\begin{center}
    {\Large\bf Intentional Electromagnetic Interference Attacks on Facial Recognition \\ 
    \vspace{0.2em} \Large Supplementary Materials}
    
    \vspace{1.5em}

    \vspace{2.5em}
\end{center}

\maketitle
\thispagestyle{empty}

\section{Test Parameters}

\begin{table}[!htbp]
\centering
\caption{Experimental Reference Table}
\label{tab:ExperimentParams}
\begin{tabular}{cccc}
\toprule
\thead{\textbf{Baseband} \\ \textbf{Frequency}} & 
\thead{\textbf{Voltage}} & 
\thead{\textbf{Modulation} \\ \textbf{Type}} & 
\thead{\textbf{Modulation} \\ \textbf{Freq.}} \\
\midrule
11.455 MHz & 2 Vpp & FM, Triangle & 190 kHz \\
\bottomrule
\end{tabular}
\end{table}

\section{Model Embedding Space Analysis}
\begin{figure}[!htbp]
     \centering
     
     \begin{subfigure}[b]{0.78\linewidth}
          \centering
          \includegraphics[width=\textwidth]{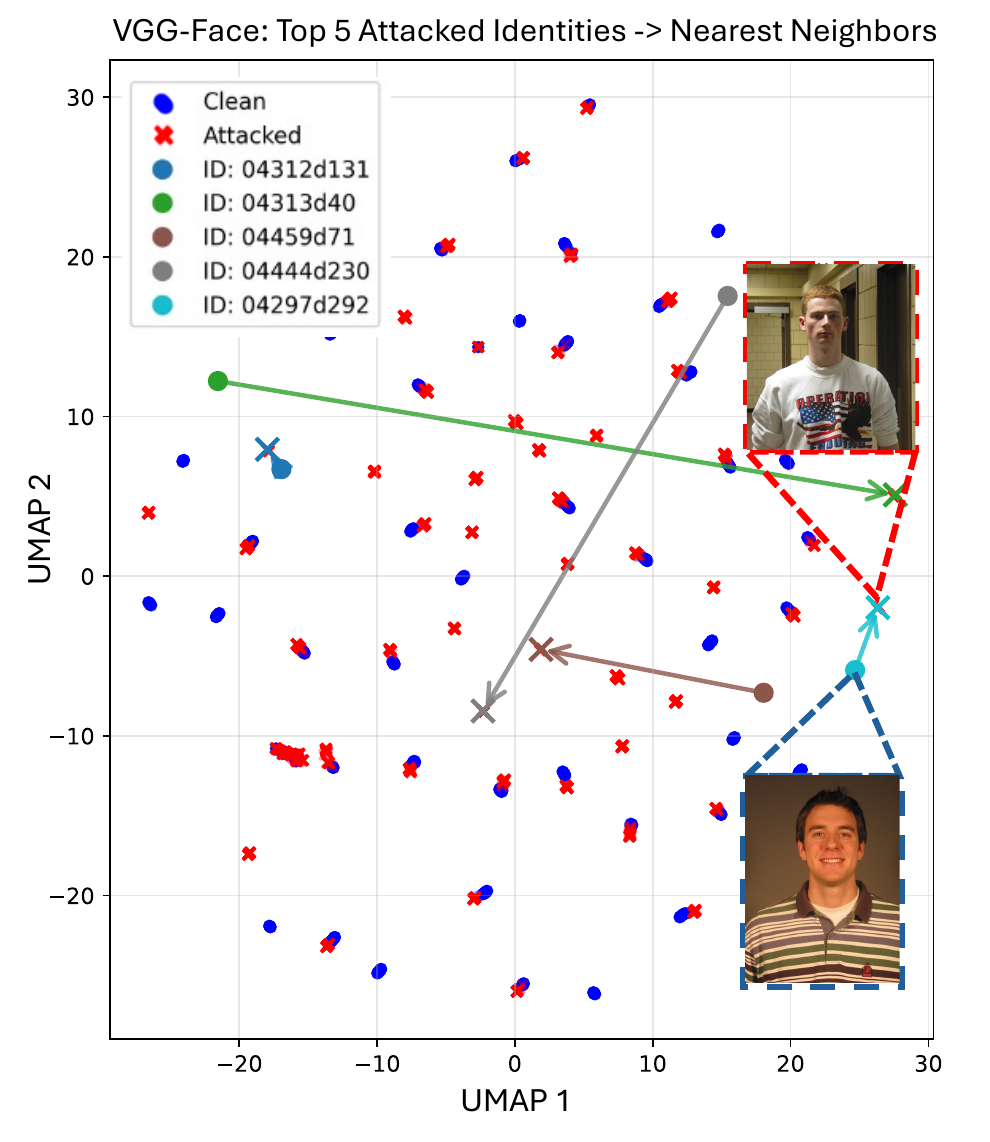} 
          \caption{Attack trajectory for the VGG-Face model.} \label{fig:traj_Vgg}
     \end{subfigure}
     \vspace{1.0em} 
     \begin{subfigure}[b]{0.78\linewidth}
          \centering
          \includegraphics[width=\textwidth]{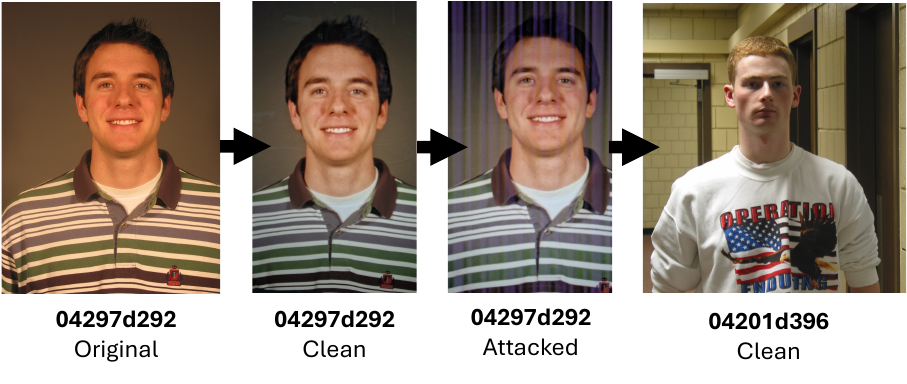} 
          \caption{Identity mapping for the VGG-Face model.} \label{fig:id_vgg}
     \end{subfigure}

     \caption{Evaluation of IEMI hardware attacks: The physical attack trajectory (top) and the respective identity transition (bottom) for VGG-Face.}
     \label{fig:FullModelStack}
\end{figure}

\begin{figure}[!htbp]
     \centering
     
     \begin{subfigure}[b]{0.78\linewidth}
          \centering
          \includegraphics[width=\textwidth]{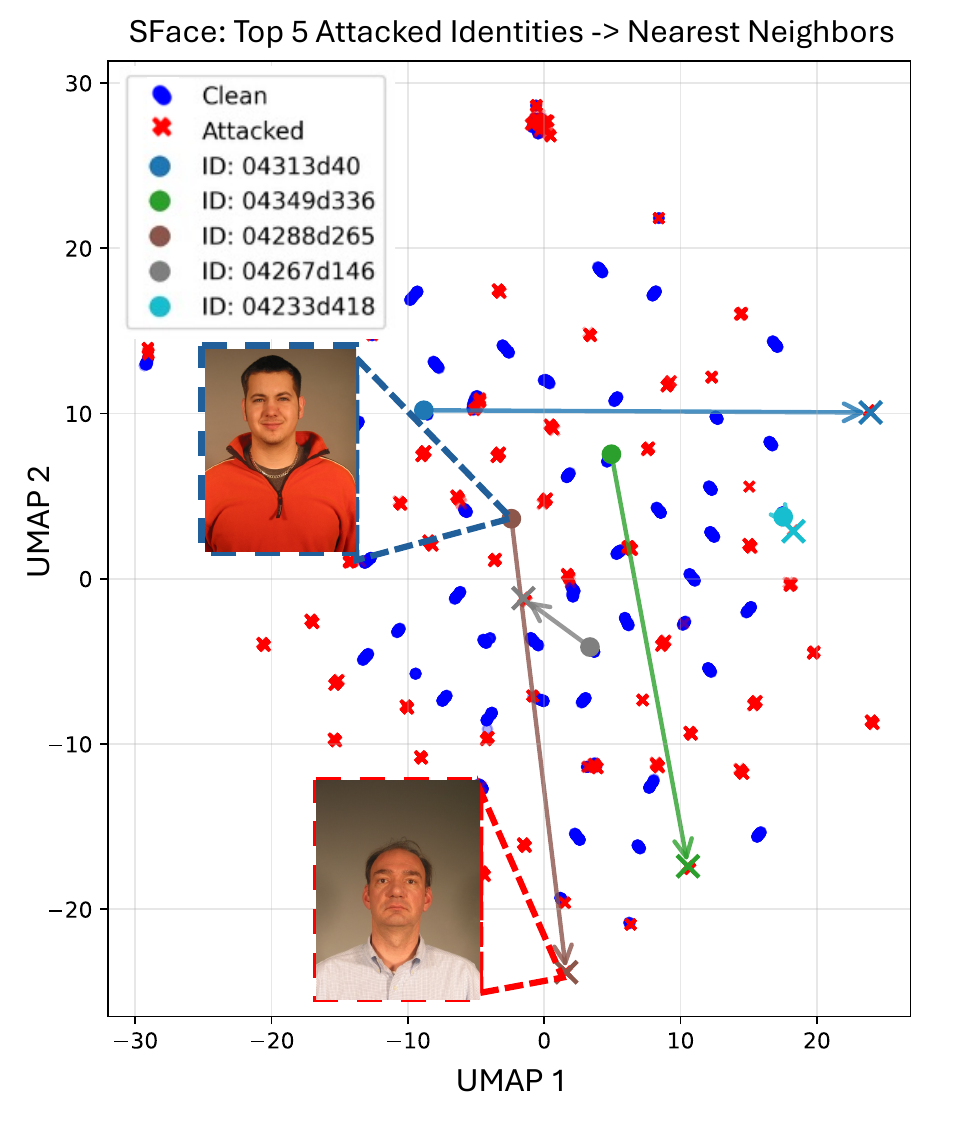}
          \caption{Attack trajectory for the SFace model.} \label{fig:traj_sface}
     \end{subfigure}
     \vspace{1.0em} 
     \begin{subfigure}[b]{0.78\linewidth}
          \centering
          \includegraphics[width=\textwidth]{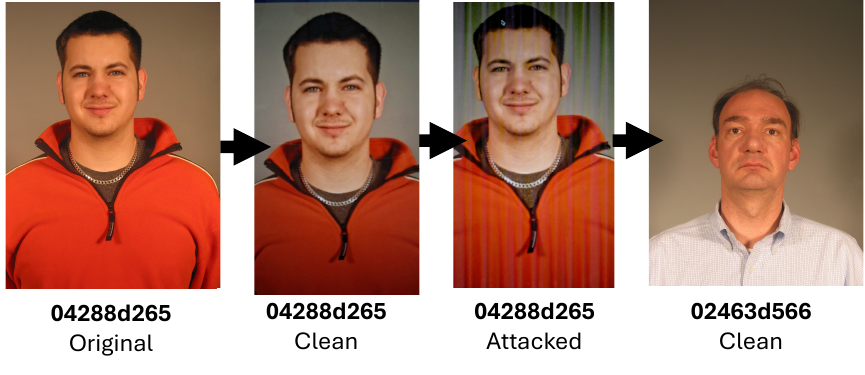}
          \caption{Identity mapping for the SFace model.} \label{fig:id_sface}
     \end{subfigure}

     \caption{Evaluation of IEMI hardware attacks: SFace (continued).}
\end{figure}

\begin{figure}[!htbp]
     \centering
     
     \begin{subfigure}[b]{0.78\linewidth}
          \centering
          \includegraphics[width=\textwidth]{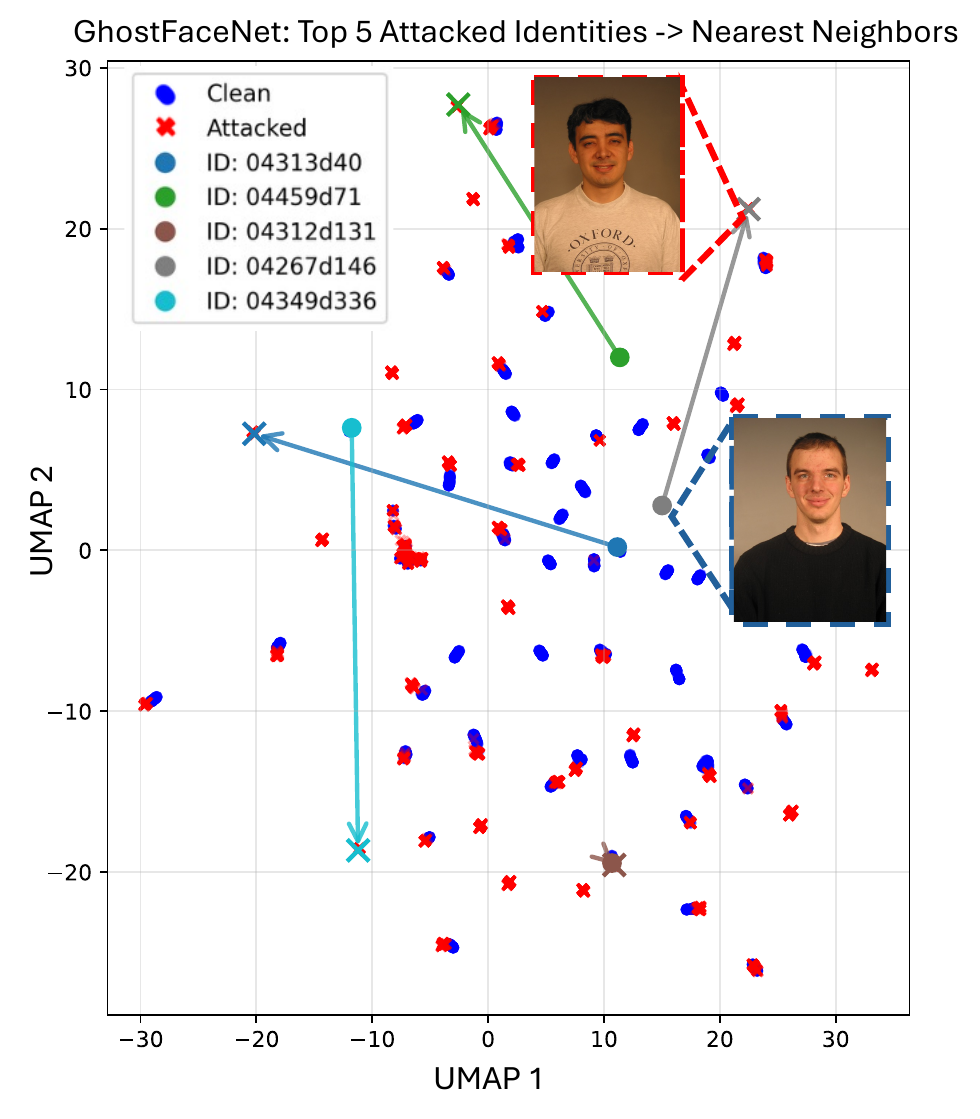}
          \caption{Attack trajectory for the GhostFaceNet model.} \label{fig:traj_3}
     \end{subfigure}
     \vspace{1.0em} 
     \begin{subfigure}[b]{0.78\linewidth}
          \centering
          \includegraphics[width=\textwidth]{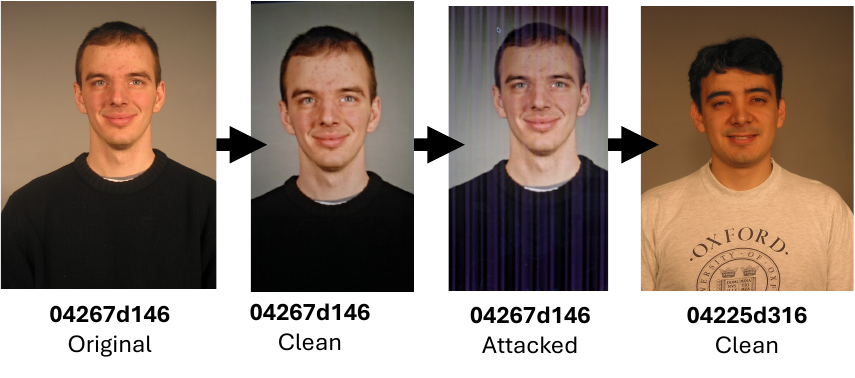}
          \caption{Identity mapping for the GhostFaceNet model.} \label{fig:id_3}
     \end{subfigure}

     \caption{Evaluation of IEMI hardware attacks: GhostFaceNet (continued).}
\end{figure}

\begin{figure}[!htbp]
     \centering
     
     \begin{subfigure}[b]{0.78\linewidth}
          \centering
          \includegraphics[width=\textwidth]{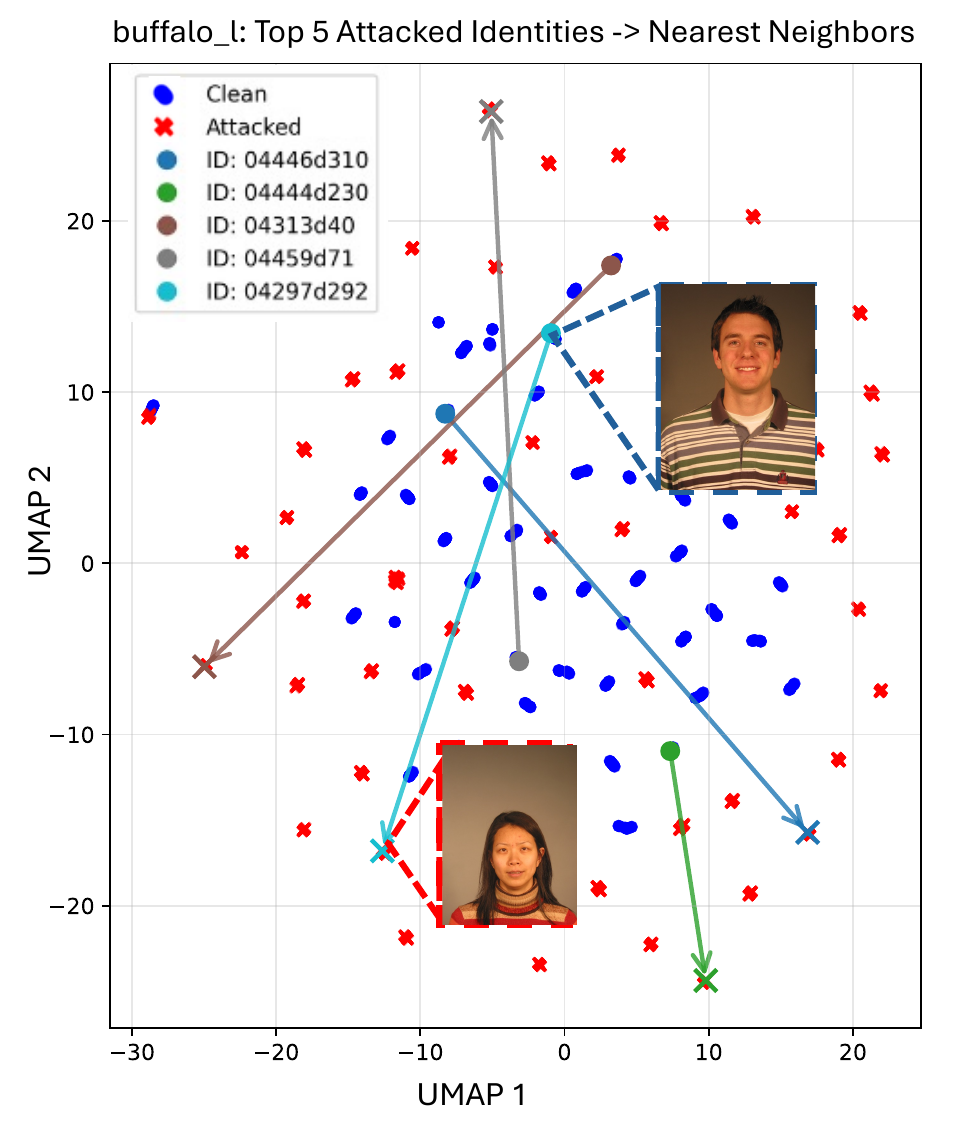}
          \caption{buffalo\_l: Attack Trajectory} \label{fig:traj_buffal0}
     \end{subfigure}
     \vspace{1.0em} 
     \begin{subfigure}[b]{0.78\linewidth}
          \centering
          \includegraphics[width=\textwidth]{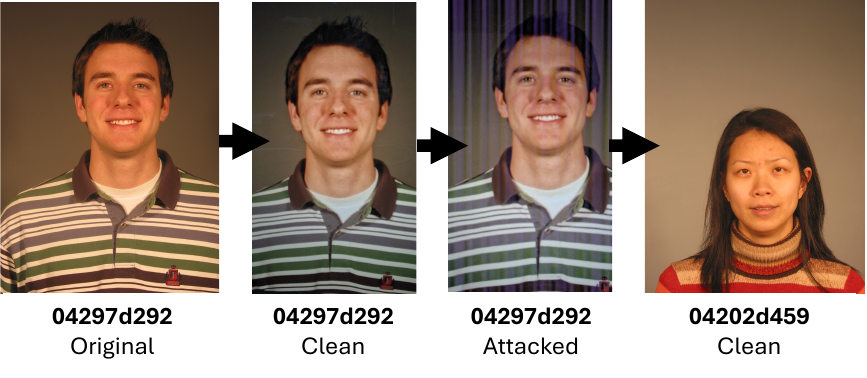}
          \caption{buffalo\_l: Identity Mapping} \label{fig:id_buffalo}
     \end{subfigure}

     \caption{Evaluation of IEMI hardware attacks: buffalo\_l (continued).}
\end{figure}

\begin{figure}[!htbp]
     \centering
     
     \begin{subfigure}[b]{0.78\linewidth}
          \centering
          \includegraphics[width=\textwidth]{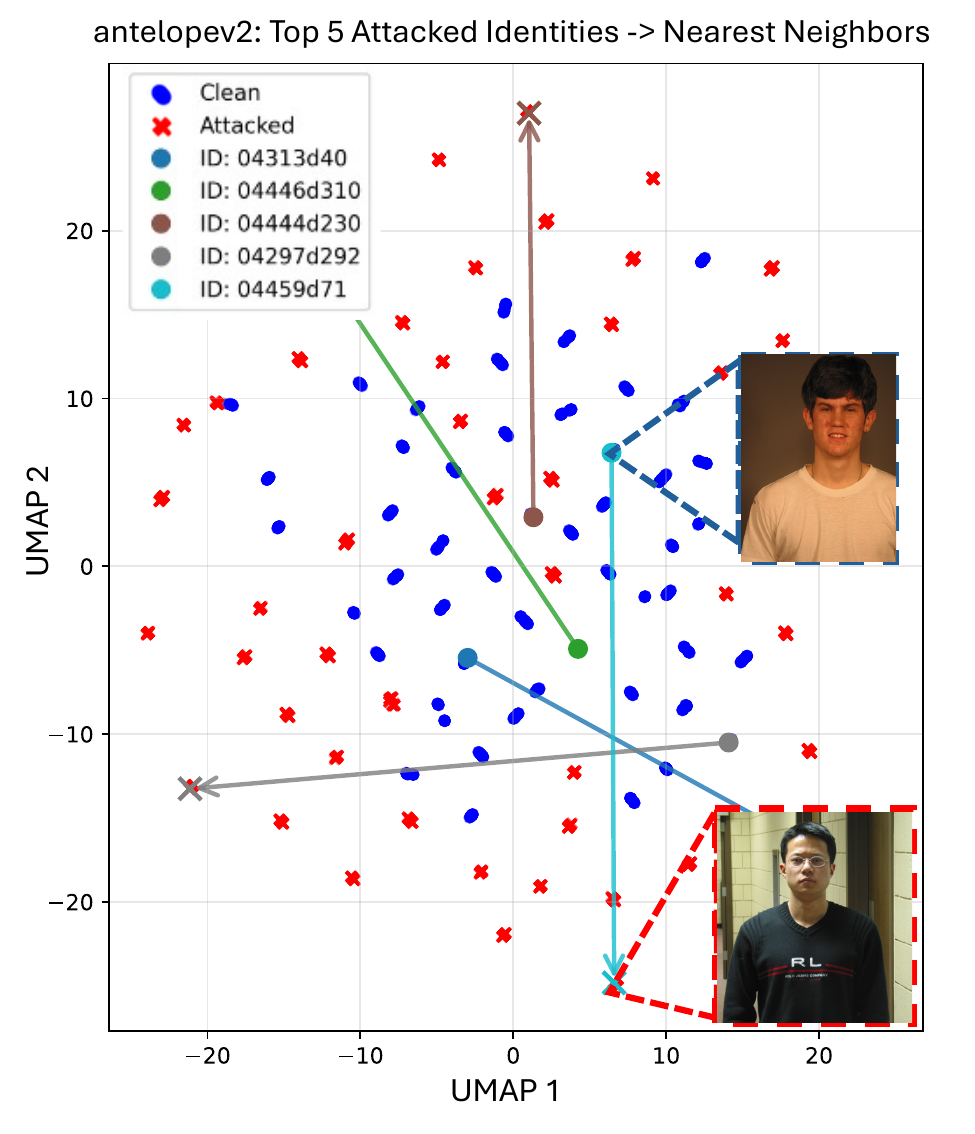}
          \caption{antelopev2: Attack Trajectory} \label{fig:traj_antelopev2}
     \end{subfigure}
     \vspace{1.0em} 
     \begin{subfigure}[b]{0.78\linewidth}
          \centering
          \includegraphics[width=\textwidth]{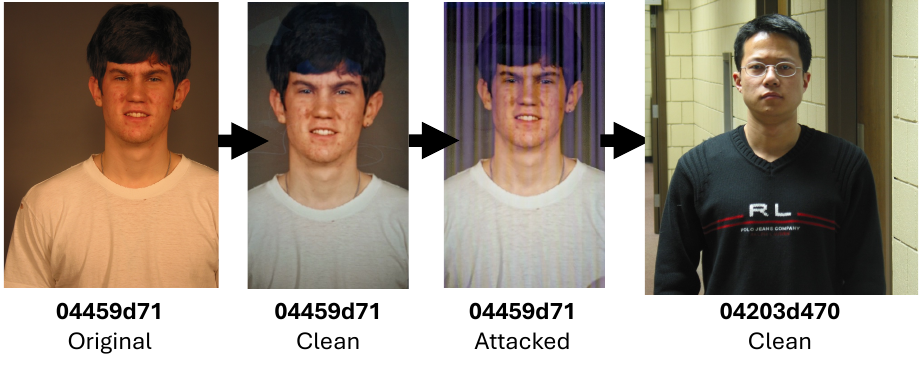}
          \caption{antelopev2: Identity Mapping} \label{fig:id_anteoplev2}
     \end{subfigure}

     \caption{Evaluation of IEMI hardware attacks: antelopev2 (continued).}
\end{figure}

\begin{figure}[!htbp]
     \centering
     
     \begin{subfigure}[b]{0.78\linewidth}
          \centering
          \includegraphics[width=\textwidth]{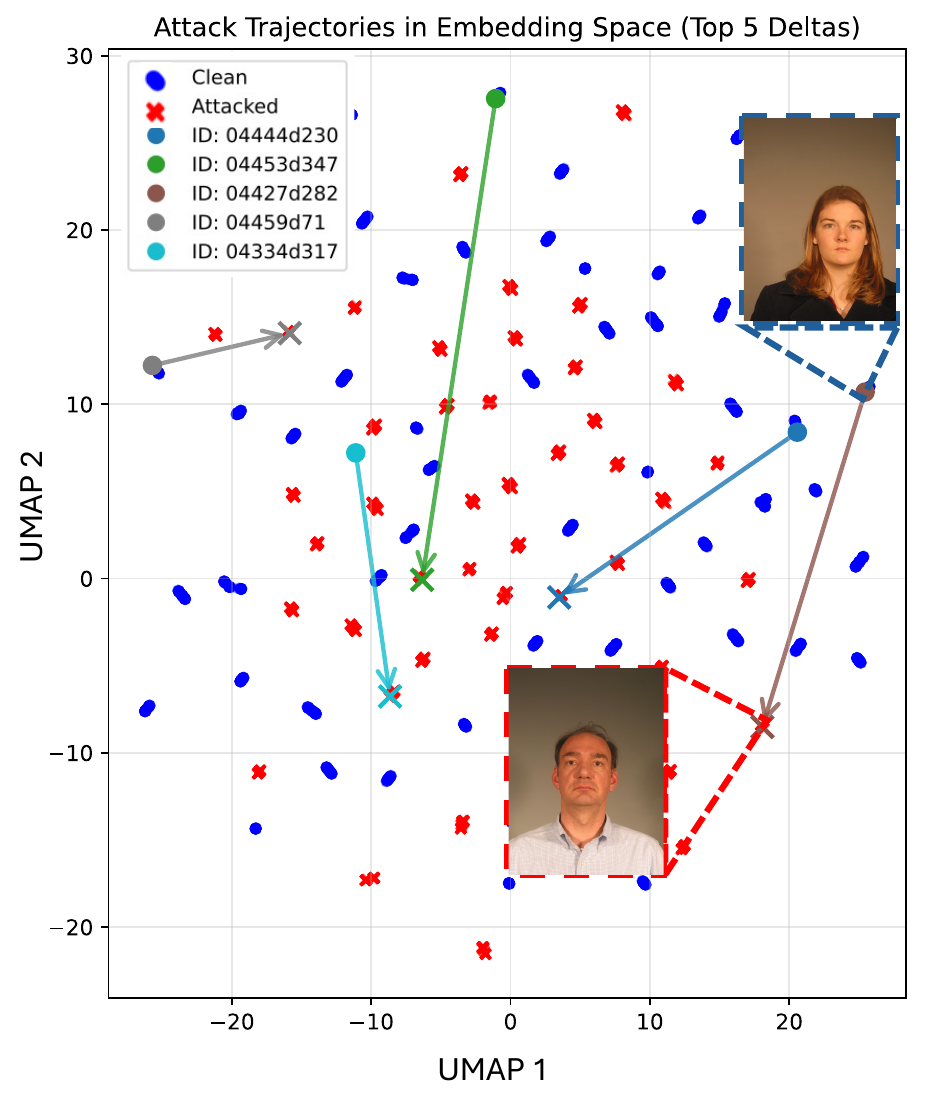}
          \caption{DINOv3: Attack Trajectory} \label{fig:traj_DINO}
     \end{subfigure}
     \vspace{1.0em} 
     \begin{subfigure}[b]{0.78\linewidth}
          \centering
          \includegraphics[width=\textwidth]{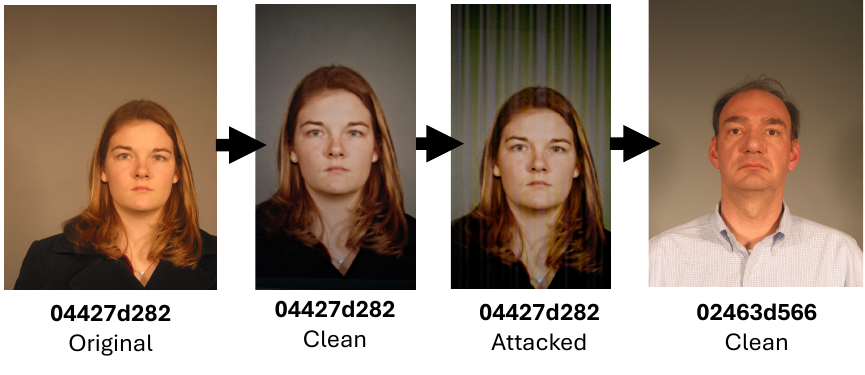}
          \caption{DINOv3: Identity Mapping} \label{fig:id_DINO}
     \end{subfigure}

     \caption{Evaluation of IEMI hardware attacks: DINOv3 (continued).}
\end{figure}

UMAP projection of the representations of top 5 attacked identities for each model tested at FMR=5.0\% \ref{fig:traj_Vgg}-\ref{fig:id_DINO}.

\clearpage 

\clearpage 

\clearpage 

\clearpage 

\clearpage 

\clearpage 

\end{document}